\documentclass{article} 
\usepackage{iclr2025_conference,times}

\usepackage{amsmath,amsfonts,bm}

\def\eqref#1{equation~\ref{#1}}

\def\1{\bm{1}}

\DeclareMathAlphabet{\mathsfit}{\encodingdefault}{\sfdefault}{m}{sl}
\SetMathAlphabet{\mathsfit}{bold}{\encodingdefault}{\sfdefault}{bx}{n}

\usepackage{hyperref}
\usepackage{url}

\usepackage[utf8]{inputenc} 
\usepackage[T1]{fontenc}    
\usepackage{booktabs}       
\usepackage{amsfonts}       
\usepackage{nicefrac}       
\usepackage{microtype}      
\usepackage{xcolor}         

\usepackage{graphicx}
\usepackage{subfig}
\usepackage{amsmath}
\usepackage{multirow}
\usepackage{booktabs}
\usepackage{algorithm}
\usepackage{algpseudocode}
\usepackage{pifont}
\usepackage[bottom]{footmisc}
\usepackage{threeparttable} 
\usepackage{makecell}
\usepackage{bm}

\title{Towards Accurate and Efficient Sub-8-Bit Integer Training}

\iclrfinalcopy
\author{{Wenjin Guo*, Donglai Liu*, Weiying Xie }\thanks{Equal contribution. }\thanks{Coresponding author. }, Yunsong Li,  Zihan Meng \\
Xidian University\\
\texttt{guowenjin@stu.xidian.edu.cn, wyxie@xidian.edu.cn} \\
\And
Xuefei Ning, Shulin Zeng, Yu Wang \\
Tsinghua University \\
\AND
Jie Lei \\
University of Technology Sydney \\
\AND
Zhenman Fang \\
Simon Fraser University
}

\begin{document}

\maketitle

\begin{abstract}
Neural network training is a memory- and compute-intensive task. Quantization, which enables low-bitwidth formats in training, can significantly mitigate the workload. To reduce quantization error, recent methods have developed new data formats and additional pre-processing operations on quantizers. However, it remains quite challenging to achieve high accuracy and efficiency simultaneously. In this paper, we explore sub-8-bit integer training from its essence of gradient descent optimization. Our integer training framework includes two components: ShiftQuant to realize accurate gradient estimation, and L1 normalization to smoothen the loss landscape. ShiftQuant attains performance that approaches the theoretical upper bound of group quantization. Furthermore, it liberates group quantization from inefficient memory rearrangement. The L1 normalization facilitates the implementation of fully quantized normalization layers with impressive convergence accuracy. Our method frees sub-8-bit integer training from pre-processing and supports general devices. This framework achieves negligible accuracy loss across various neural networks and tasks ($0.92\%$ on 4-bit ResNets, $0.61\%$ on 6-bit Transformers). The prototypical implementation of ShiftQuant achieves more than $1.85\times/15.3\%$ performance improvement on CPU/GPU compared to its FP16 counterparts, and $33.9\%$ resource consumption reduction on FPGA than the FP16 counterparts. The proposed fully-quantized L1 normalization layers achieve more than $35.54\%$ improvement in throughout on CPU compared to traditional L2 normalization layers. Moreover, theoretical analysis verifies the advancement of our method. 
\end{abstract}

\section{Introduction}
Recently, deep neural networks have made significant advancements in various tasks. However, this progress comes at the cost of high computational complexity. High energy cost and computational resource consumption hamper the deployment of these deep learning applications on both edge-device and high-end cloud servers. While existing studies primarily concentrate on minimizing resource consumption during inference \cite{inference1, inference2, inference3, inference4, inference5, inference6, inference7}, it should be noted that the computational complexity of training is approximately three times that of inference \cite{banner2018scalable}. Consequently, there is an urgent demand for efficient training methods.

Reducing the bitwidth of data is a reasonable approach. The resource consumption of a neural network scales almost linearly with the bitwidth of data \cite{bitwidth}. However, low-precision training faces two challenges. The first is the wide range of gradients. Very few outliers extremely expand the data range, resulting in high quantization error \cite{sun2020ultra, xi2023training, chen2020statistical}. The second is the sharp loss landscape of low-precision networks \cite{low-precision-landscape}. Sharp local minimal points disrupt optimization significantly \cite{sharpness-generality1, visualizing-landscape}.  

To address the challenge of wide gradient range and significant quantization errors, common approaches include (1) employing a non-uniform quantization format \cite{sun2020ultra, cambier2020shifted, zhong2022exploring}, (2) utilizing smaller quantization granularity, or (3) suppressing outliers before quantization \cite{xi2023training, chen2020statistical}. However, all methods encounter practicality issues, as they cannot be easily adapted to existing General Matrix Multiply (GEMM) software and hardware implementations or extremely exacerbate the computation burden (detailed discussion in Sec. \ref{3}). To this end, we analyze the structure of outliers in gradients and develop an efficient and effective quantizer, ShiftQuant. With strategic and structural grouping of channels, ShiftQuant achieves excellent outlier suppression at a negligible cost, and also supports GEMM. 

Recent methods focus on developing new quantizers but neglect the impact of sharp loss landscape \cite{xi2023training,chen2020statistical,sun2020ultra,fu2021cpt,wu2022drgs,das2018mixed,banner2018scalable,luq,ghaffari2022integer}. The sharp landscape of low-precision networks brings more local minimal points and leads to unstable convergence. Moreover, sharp curvature demands more bits to specify gradients \cite{dong2019hawq, sharpness-generality1}. In this paper, we reveal that the source of the sharp landscape lies in the limited smoothening and quantization-tolerated capacity of the normalization layers \cite{batchnorm, layernorm, instancenorm, groupnorm}. Stronger smoothening comes from stronger regularization. In this paper, we introduce the stronger regularization, L1 normalization, into normalization layers, which achieves clearly smooth loss landscape under less computation. 

Our main contributions include the following: 

By analyzing the structure of outliers in gradients, we find that appropriately grouping channels can effectively reduce quantization errors. Hence, we develop a new quantizer, ShiftQuant. ShiftQuant applies a smart grouping strategy that effectively approximates the optimal solution for grouping with minimal computational overhead. ShiftQuant utilizes the common low-bitwidth integer format and supports GEMM. Moreover, we develop a specific implementation of the new quantizer, which avoids the inefficient memory rearrangement in per-group quantization for the first time. 

We experimentally analyze the source of the sharp loss landscape in low-precision networks. Quantization weakens the smoothening effort of L2 normalization layers significantly. To this end, we develop L1 normalization layers, which achieves stronger smoothening with less computation. 

We evaluate the proposed framework on various datasets. Our method achieves negligible performance loss on ResNets, Transformers, GNN, and RNN under sub-8-bit integer training, respectively. Furthermore, theoretical analysis substantiates the effectiveness of the new quantizer and normalization layers. 

Our prototypical implementation of ShiftQuant achieves more than $1.85\times/15.3\%$ performance improvement on CPU (ARMv8)/GPU (Nvidia RTX 3090) compared to Pytorch.fp16, and more than $33.9\%$ resource consumption reduction on FPGA (ZC706) compared to FP16. Moreover, 6-bit ShiftQuant surpasses 4-bit integer training competitor in efficiency. The implementation of proposed fully-quantized L1 normalization layers achieves more than $35.54\%$ improvement of throughout on CPU compared to traditional L2 normalization layers. 

\section{Related Work and Challenges}\label{3}
\subsection{New Quantization Data Formats}
Developing new data formats with wide representing range reduces quantization loss significantly. Sun et al. \cite{sun2020ultra} proposes a new fixed-point format using radix-4, which, unlike the traditional radix-2 fixed-point format, offers a significantly larger representation range and higher resolution. Fu et al. \cite{fu2021cpt} explores a dynamic data format with varying bit-widths to balance accuracy and efficiency by periodically adjusting the bit-width. Cambier et al.  \cite{cambier2020shifted} proposes the S2FP8 format, where a vector is represented by an FP8 vector and two FP32 values, named squeeze and shift factors. Zhong et al. \cite{zhong2022exploring} develops the MLS tensor format, which balances the accuracy and computation efficiency in a tensor level. Although the excellent accuracy, new data formats demand customer design of new hardware units, hindering the deployment on contemporary devices inherently. 
\subsection{Fine-grained Quantization}
Fine-grained quantization can reduce quantization error significantly. However, naive fine-grained quantization may lead to incompatibility with GEMM \cite{smoothquant,zhong2022exploring}. Since when fine-grained quantization is applied to the inner dimension, different indices have distinct floating-point scaling factors. Inevitably, when accumulating these indices, each element must first multiply its corresponding scaling factor, thereby introducing extra floating-point multiplication operations. As shown in Figure \ref{MMs}, three matrix multiplies are involved in training: forward propagation, loss propagation, and calculation of weight gradient. Almost all dimensions serve as the inner dimension. To apply GEMM, fine-grained quantization is not allowed. 
\subsection{Outlier Suppression before Quantization}
Before quantization, the data is reflected to an easy-quantization space through an invertible mapping. By conducting the inverse mapping after dequantization, the original data can be recovered with small loss. Chen et al. \cite{chen2020statistical} implements the reflection by multiplying gradients with a constructed Hadamard matrix. Similarly, Xi et al. \cite{xi2023training} applies the Hadamard reflection on quantizing activation during training transformers. Additional computation arises from constructing reflection. Moreover, the reflection operation is conducted in expensive floating-point format. 

\begin{figure}[]
	\centering
	\begin{minipage}{0.30\textwidth}
		\includegraphics[width=\textwidth]{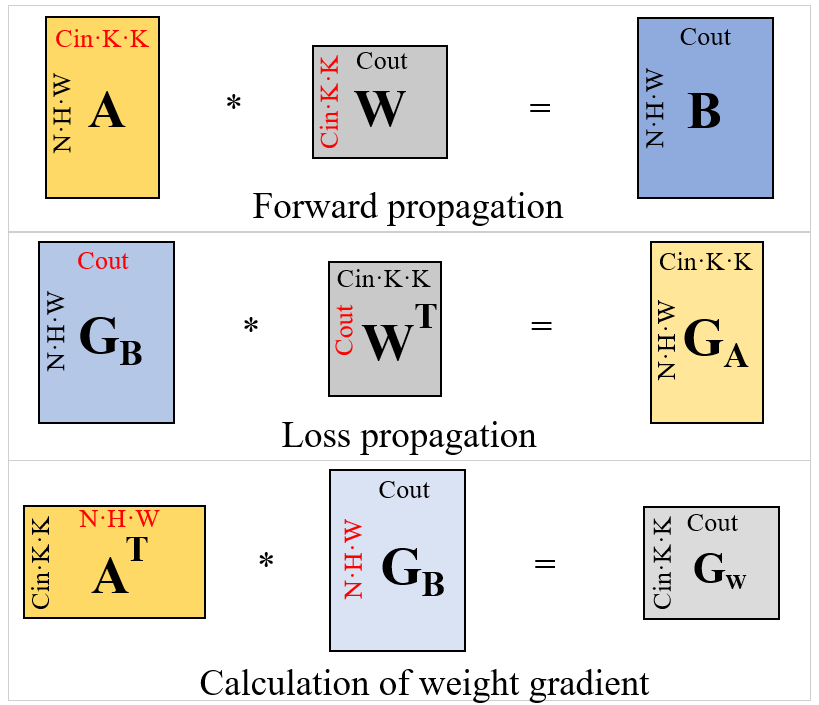}
		\caption{Matrix multiplications in training. Inner dimensions are labeled in red. }
		\label{MMs}
	\end{minipage}
	\hfill
	\begin{minipage}{0.68\textwidth}
		\scriptsize
		\centering
		\vspace{-15pt}
		\renewcommand{\arraystretch}{1.3}
		\setlength{\tabcolsep}{1.0pt}
		\captionof{table}{Difference between our method and prior studies. }
		\begin{tabular}{l|c|c|c|c}
			\toprule
			& \multicolumn{1}{l|}{\makecell{New data format\\ \cite{sun2020ultra},\\ \cite{fu2021cpt},\\ \cite{cambier2020shifted}}} & \multicolumn{1}{l|}{\makecell{Fine-grained\\ \cite{das2018mixed},\\ \cite{wu2022drgs}}} & \multicolumn{1}{l|}{\makecell{Outlier suppression \\before quantization\\ \cite{chen2020statistical},\\ \cite{xi2023training}}}  & Ours \\ \hline
			General data format                  &\textcolor{red}{\ding{56}}   & \textcolor{green}{\ding{52}} & \textcolor{green}{\ding{52}} & \textcolor{green}{\ding{52}} \\ \hline
			GEMM supporting                      & -                           & \textcolor{red}{\ding{56}}   & \textcolor{green}{\ding{52}} & \textcolor{green}{\ding{52}} \\ \hline
			No pre-processing   & \multirow{2}{*}{\textcolor{green}{\ding{52}}}& \multirow{2}{*}{\textcolor{green}{\ding{52}}}&\textcolor{red}{\ding{56}}& \multirow{2}{*}{\textcolor{green}{\ding{52}}} \\
			on quantization     &      &     &  &     \\ \hline
			Fully-quantized      & \textcolor{red}{\ding{56}}&\textcolor{red}{\ding{56}}&\textcolor{red}{\ding{56}}&\multirow{2}{*}{\textcolor{green}{\ding{52}}} \\
			normalization layers &   &     & &   \\ \bottomrule
		\end{tabular}
		\label{contribution}
	\end{minipage}
\end{figure}

Our method utilizes the common integer format. Meanwhile, no pre-processing is applied before quantization. We highlight the difference between our method and prior methods at Table  \ref{contribution}. 
\section{ShiftQuant}
\begin{figure}[h]
	\centering 
	\includegraphics[width=1\linewidth, height=0.4\linewidth]{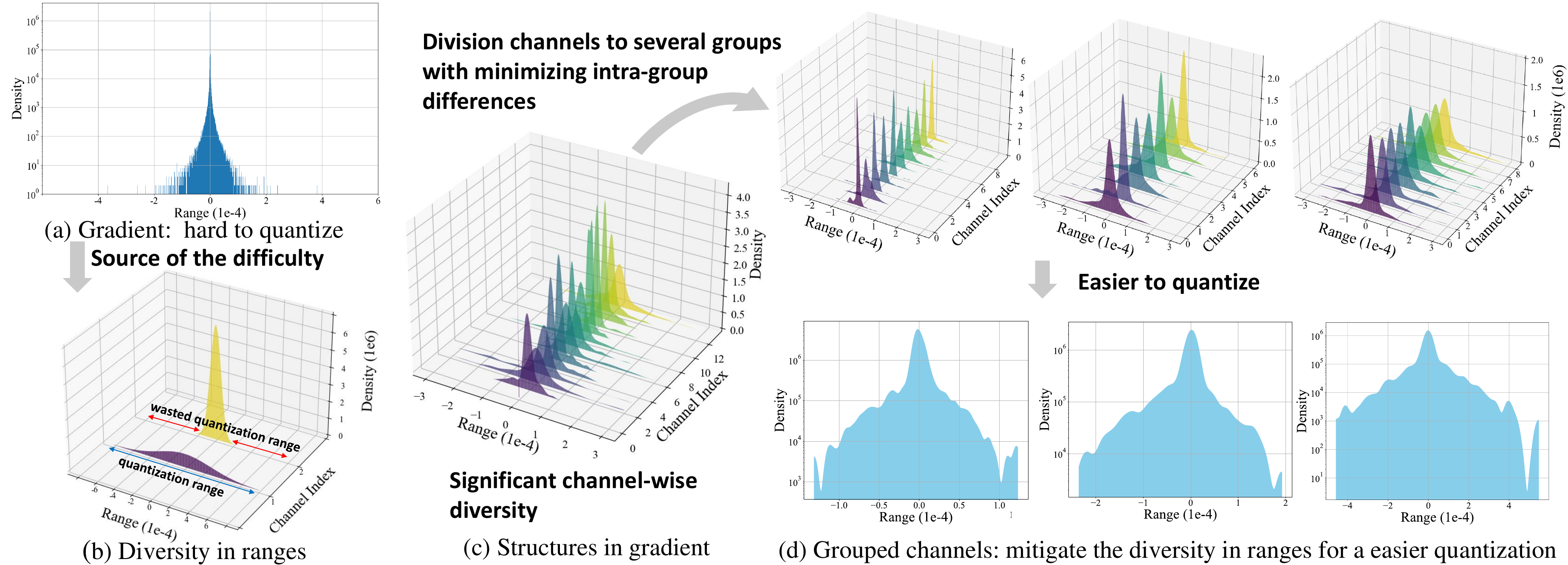}
	\caption{The difficulty of quantizing gradients comes from the diversity between channels. ShiftQuant aims to minimize the diversity through strategic grouping channels. (a) Gradients exhibit extremely bell-curve distribution, which is very difficult to quantize. (b) Diversity in magnitude of channels leads to high quantization error. (c) High diversity of channels in gradients. (d) ShiftQuant divides channels  to several groups. Each group characters small diversity, leading to easier quantization. }
	\label{meth}
\end{figure}
\subsection{Smart Channel Grouping}

As shown in Figure \ref{meth}(c), the distribution of gradient varies among channel dimension extremely. Channels with wide range expand the quantization range, leading to extremely fewer quantization levels to represent channels with small range (as shown in Figure \ref{meth}(b)). High quantization variance comes from the diversity in channels' ranges. Merging the gap between channel's range can reduce quantization error significantly. Grouping channels is a reasonable approach. Appropriately grouping channels can minimize the diversity in each group. Then, per-group quantization can achieve low quantization variance and high efficiency at the same time. 

\textbf{Optimizing grouping strategy}

It is intuitive that smaller channels should be divided to one group, and larger channels should be divided to another group. The key of the grouping strategy is to find the threshold to distinguish small and large channels. To divide channels into $N_G$ groups, we identify $N_G+1$ thresholds $\tau_0, \tau_1, \cdots, \tau_{N_{G}-1}, \tau_{N_{G}}$. 

The object of grouping is: 
\begin{equation} \label{dq_f}
	\begin{aligned}
		\min_{\tau_0, \tau_1, \cdots, \tau_{N_{G}-1}, \tau_{N_{G}}} & \sum_{g=1}^{N_{G}} \sum_{i \in P^g} \frac{\tau_g}{r_{i}}, \\
		s.t.  \, r_{max} = \tau_0 \ge & \tau_1 \ge \cdots \ge \tau_{N_{G}-1} \ge \tau_{N_{G}} = 0 , 
	\end{aligned}
\end{equation}
where $P^g$ is a set that contains the indexes of channels allotted to group $g$. $r_{max}$ is the minimum around ranges of channels. $r_i$ is the range of $i$-th channel. $r_i$ satisfies the following condition:
\begin{equation}
	\begin{aligned}
		\tau_{g-1} < r_i \leq \tau_{g},\, if \, i\in P^g. 
	\end{aligned}
\end{equation}
$\sum_{i \in P^g} \frac{\tau_g}{r_i}$ reflects the diversity of channels' range in group $g$. Moreover, problem (\ref{dq_f}) is equivalent to directly optimize the quantization variance. The theoretical verification can be found in Appendix. \ref{equal}. 

\textbf{Efficient power-of-two grouping}

We can apply the optimal solution of problem (\ref{dq_f}) to group channels. Unfortunately, two serious obstacles will rise. The first is the complexity to solve. Problem (\ref{dq_f}) does not have a closed-form solution. It requires nonlinear dynamic programming, which demands iterations of complicated calculation to find the optimal solution. The second is the optimized thresholds are in floating-point formats. The scale of these groups does not follow integer-times relation. We have to accumulate the results of groups in floating-point format. 

We analyze the characteristics of the objective function $\frac{\tau}{r}$, the ratio of group's upper threshold to the range of channel. We find that larger threshold can tolerate greater distance between channels' range and threshold. For instance, let $\tau_p=2\tau_q, r_i=2r_j$. The loss incured by grouping channel $i$ to group $p$ is equal to that of grouping channel $j$ to group $q$ ($\frac{\tau_p}{r_i}=\frac{\tau_q}{r_j}$). However, the distance between the threshold and channel's range of group $p$ is twice of group $q$ ($\tau_p-r_i=2(\tau_q-r_j)$). Therefore, to minimize the objective function, a group with larger threshold should hold larger range.  

Consider the accuracy and hardware efficiency, we impose power-of-two relation on thresholds: 
\begin{equation}
	\tau_{g} = \tau_{0} \cdot 2^{g} = r_{max} \cdot 2^{g}, g=1, \cdots, G. 
\end{equation}
The above strategy partitions more ranges for groups with larger threshold. Moreover, it frees grouping from costly optimization and enables accumulation groups in integer format with only a shift operation. Take the calculation of $\bm{G_{A}}$ as example: 
\begin{equation}\label{dq_mm_f}
	\begin{aligned}
		\bm{G}_{A} = \bm{G}_{B} \cdot \bm{W}^{T} & = \sum_{g=1}^{N_{G}} [s^{-1}_{\bm{G}_{B}} \cdot 2^{-g} \cdot {(\bm{G}_{B})}^g_q] \cdot [s^{-1}_{\bm{W}} \cdot {(\bm{W}^T)}^g_q] \\
		& = s^{-1}_{\bm{G}_{B}} \cdot s^{-1}_{\bm{W}} \sum_{g=1}^{N_{G}}[{(\bm{G}_{B})}^g_q \cdot {(\bm{W}^T)}^g_q >> g], 
	\end{aligned}
\end{equation}
where $(\bm{G}_{B})^g_q$ and $(\bm{W}^T)^g_q$ denote the quantized $g$-th group of $\bm{G}_{B}$ and $\bm{W}^{T}$. Eq. (\ref{dq_mm_f}) is also diagrammed in Figure \ref{imple}(b). All of the calculation are performed on integer format. Detailed derivation is provided in Appendix. \ref{shiftMM}. ShiftQuant achieves unbiased and low-variance quantization. Theoretical analysis is provided in  Appendix. \ref{anashiftquant}. 

\subsection{Implementation Optimization}
We devise two approaches to implement ShiftQuant. One is based on GEMM. Additionally, we develop a specific implementation method, named ShiftMM, which achieves higher hardware efficiency. 
\begin{figure}[h]
	\centering 
	\includegraphics[width=0.95\linewidth, height=0.45\linewidth]{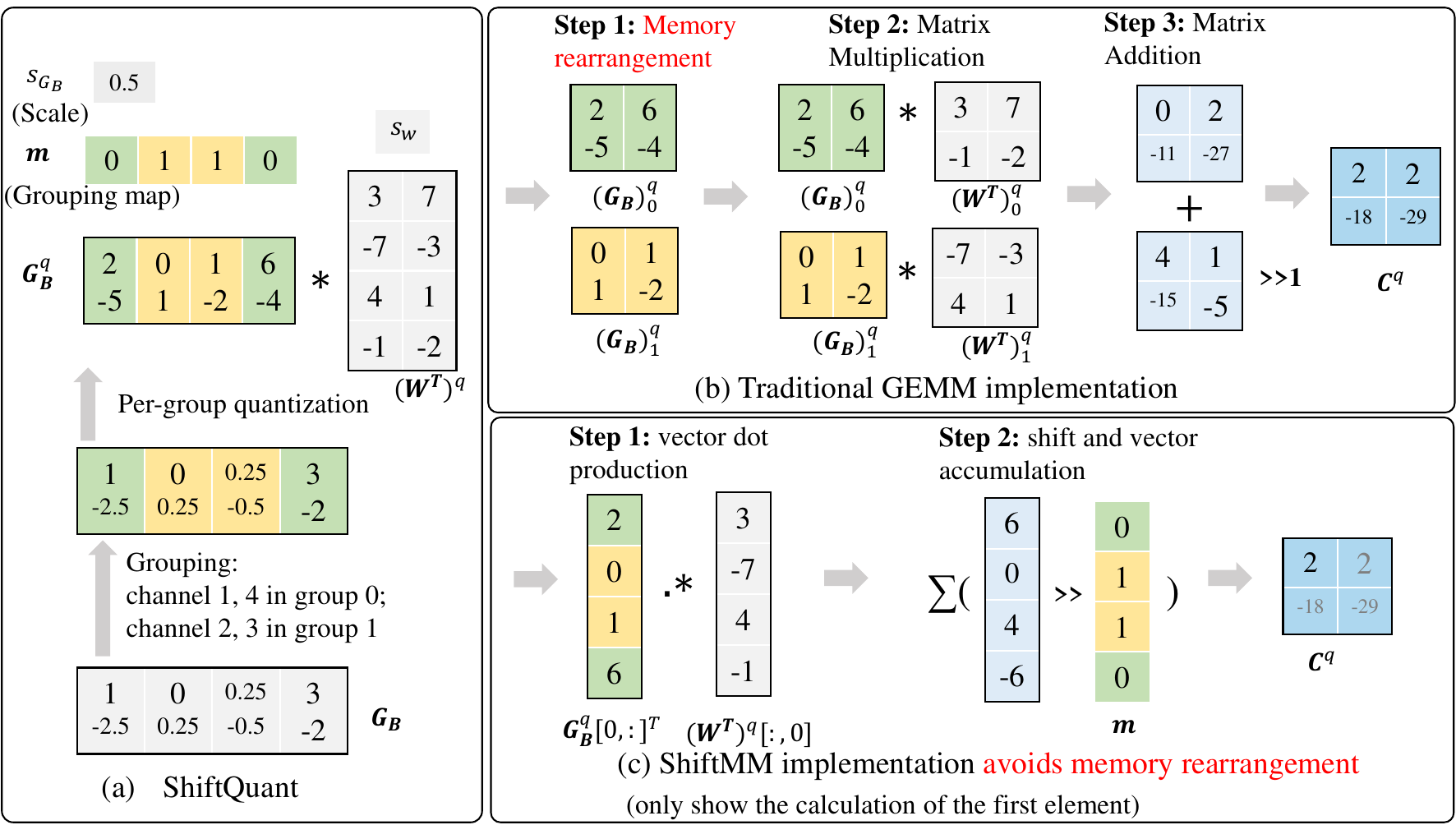}
	\caption{The power-of-two grouping strategy in ShiftQuant paves a way for implementing per-grouping quantization without memory rearrangement. \textbf{Traditional implementation} approach (b) focuses on applying off-the-peg packages. It transfers original matrix multiplication to several small matrix multiplications based on the grouping map $\bm{m}$, which involves expensive memory rearrangement.  \textbf{ShiftMM} (c) aims higher hardware efficiency. It replaces the memory rearrangement with only a low-cost shift operation. Meanwhile, it only demands slight changes on off-the-peg packages. }
	\label{imple}
\end{figure}

\textbf{Implemented by GEMM}

As illustrated in Figure \ref{imple}(b), ShiftQuant divides the traditional matrix multiplication into $n$ pairs of smaller matrix multiplications. Implementing ShiftQuant through GEMM involves three steps. Firstly, extracting the $n$ pairs of smaller matrices, which requires memory rearrangement. Secondly, employing GEMM to perform the $n$ pairs of matrix multiplications. Finally, shifting and summing the results of these $n$ multiplications. This approach achieves a good balance between efficiency and versatility. However, the memory rearrangement increases time consumption, and the smaller matrix multiplications do not fully capitalize on the advantages of GEMM.

\textbf{Implemented without memory rearrangement}

ShiftQuant essentially assigns different shift amounts to each index of the inner dimension. As depicted in Figure \ref{imple}(c), we can perform the shifting directly before the accumulation phase of the vector multiplication. This technique offers two benefits. Firstly, it avoids memory rearrangement. Secondly, it maintains the same level of parallelism as the original matrix multiplication. We refer to this implementation method as ShiftMM. The implementation of ShiftMM just requires adding a shifting operation after the multiplication step in the standard GEMM code. 
\begin{figure}[h]
	\centering 
	\includegraphics[width=1\linewidth, height=0.22\linewidth]{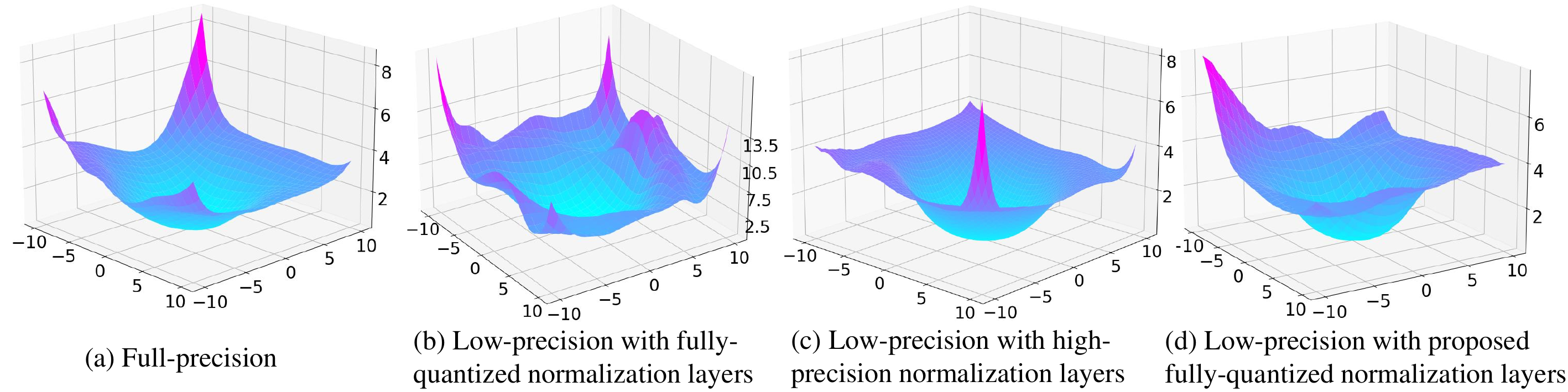}
	\caption{Low-precision networks feature sharpen loss landscape, which disrupts convergence. We visualize the  loss landscape of ResNet20 on CIFAR10 by the method in \cite{visualizing-landscape}. }
	\label{landscapes}
\end{figure}
\section{Fully-quantized L1 Normalization Layers}
\subsection{Observation and Proposal of Normalization Layers}
\begin{figure}
	\centering 
	\includegraphics[width=\linewidth, height=0.2\linewidth]{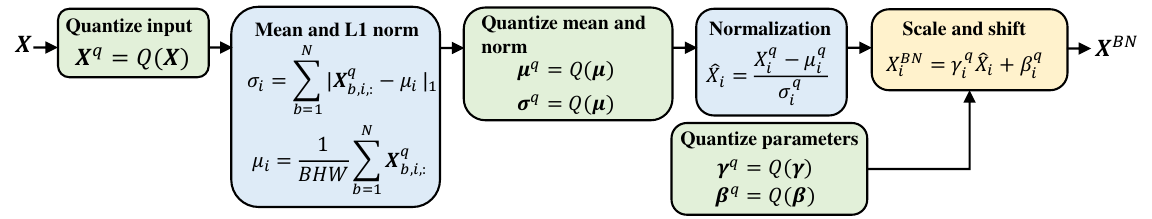}
	\caption{Procedures of the proposed fully-quantized normalization layer (take BatchNorm for example). The input $\bm{X}\in \mathcal{R}^{N\times C \times HW}$,  where $N,C,HW$ denote the batchsize, number of channels, number of spatial elements, respectively. }
	\label{l1bnq}
\end{figure}
As shown in Figure \ref{landscapes}, the loss landscape of low-precision networks is sharper and more challenging to optimize compared to that of full-precision networks. As existing literature \cite{santurkar2018does} points out that normalization layer help smoothen the loss landscape, we hypothesize that the quantization of the normalization layers might be a main cause of the sharp loss landscape of low-precision networks. To verify this hypothesis, we replace the quantized normalization layers in the low-precision networks with full-precision ones. As shown in Figure \ref{landscapes}(c), the sharpness of the loss landscape disappeared, indicating that the quantization of the L2 normalization layers diminishes its smoothening capability, leading to the sharp loss landscape. 

Two approaches can mitigate this problem, enhancing the smoothening ability of normalization layers and enhancing their tolerance to quantization. We apply L1 normalization in our fully-quantized normalization layers. Firstly, L1 normalization has stronger regularization effort than L2 normalization \cite{huber1996robust, candes2006robust}, leading to stronger smoothening ability (see Sec. \ref{l1ta}). Secondly, the L1 norm of activation are usually larger than its L2 norm. Larger L1 norm can tolerate more quantization errors than L2 norm in the processing of normalization (see Appendix. \ref{qtl1}). 

The flow of the fully-quantized L1 normalization layer is shown in Figure \ref{l1bnq}. All of the input features, statistics, and parameters are quantized to low bitwidth formats. As shown in Figure \ref{landscapes}(d), our fully-quantized L1 normalization layer achieves similar smoothening effort as the full-precision L2 normalization layer. 
\subsection{Theoretical Analysis}\label{l1ta}
The Lipschitzness constant reflects the smoothness of loss landscape \cite{boyd2004convex, bottou2018optimization}. The Lipschitzness of L1 normalization layers $L^{1}$ and L2 normalization layers $L^{2}$ satisfies the following relationship: 
\begin{equation}
	\frac{L^{1}}{L^{2}} \le \frac{\lVert \bm{x} \rVert_{2}}{\lVert \bm{x} \rVert_{1}} \le 1. 
\end{equation}
$\bm{x}$ is a vectorized input activation of normalization layers. A smaller Lipschitzness constant implies that the gradients of the loss function do not change drastically with small changes in input, which means the loss landscape is smoother \cite{boyd2004convex, bottou2018optimization}. Detailed proof can be found in Appendix. \ref{lipl1}. 
\section{Experiments}
\subsection{Performance Analysis}
We evaluate our integer training framework on ResNets \cite{he2016deep} and Transformers \cite{Vaswani2017AttentionIA, dosovitskiy2020image}. We also report the performance of recent state-of-art methods, including new data format methods CPT \cite{fu2021cpt}, Ultra-Low \cite{sun2020ultra}, reflection methods StateQuant \cite{chen2020statistical}, \cite{xi2023training}, and full integer training architecture \cite{ghaffari2022integer}. We apply ShiftQuant on activations and gradients, per-out-channel quantization on weights during forward propagation, and per-input-channel quantization on weights during loss propagation. We set the number of groups in ShiftQuant as 4 as default. We build two baselines. One is training in full-precision (fp32). The other is training in 8-bit integer (per-tensor quantization on weights, activations, and gradients). We adopt hyper-parameters, optimizers, and schedulers for all the evaluated models.

\textbf{Image classification}
We select various network architectures and perform experiments on CIFAR10 \cite{krizhevsky2009learning} and ImageNet \cite{deng2009imagenet} to validate the effectiveness of our method. In convolutional neural networks, we employ 4-bit weights, activations, and gradients, along with 8-bit fully-quantized L1 Batch Normalization (L1BNQ) on ResNets \cite{he2016deep}. As shown in Table \ref{resnet}, from the shallow ResNet18 to the deeper ResNet56, our method achieves accuracy comparable to that of the floating-point setup. For the fine-tuning tasks in Vision Transformers (ViT) \footnote{\href{https://github.com/jeonsworld/ViT-pytorch}{https://github.com/jeonsworld/ViT-pytorch}} \cite{dosovitskiy2020image}, we also use 4-bit weights, activations, and gradients. Since ViT pre-training utilized L2 Layer Normalization, we have to keep full-precision L2 layer normalization. It is a future work to quantize the L2 normalization layers of pre-trained models. As indicated in Table \ref{resnet}, our approach achieves similar performance to the floating-point models in both the basic ViT-B and the expanded ViT-L models with negligible loss of accuracy. The experiments confirm the robustness of our method across different network architectures.
\begin{table}[]  
	\centering
	\renewcommand{\arraystretch}{1.0}
	\scriptsize
	\caption{Results on ResNets. }
	\begin{tabular}{llllllll}
		\toprule
		Dataset                   & Model    & \multicolumn{2}{l}{Baselines} & CPT                               & StateQuant                                  & Ultra-low                            & Ours           \\
		&          &               &               & \cite{fu2021cpt} & \cite{chen2020statistical} & \cite{sun2020ultra} &                \\ \cline{3-8} 
		&          & FP            & INT8          & INT 3-6/3-6/6                     & INT 8/8/5                                   & radix-4 4/4/4                        & INT 4/4/4      \\ \hline
		\multirow{5}{*}{CIFAR10}  & ResNet20 & 91.25         & 87.97         & 90.13                             & -                                           & -                                    & \textbf{90.41} \\
		& ResNet38 & 92.04         & 89.39         & 91.24                             & -                                           & -                                    & \textbf{91.46} \\
		& ResNet56 & 93.03         & 88.82         & 91.97                             & -                                           & -                                    & \textbf{92.03} \\
		& ResNet18 & 93.31         & 91.73         & 92.74                             & 92.85                                       & \textbf{93.76}                       & 92.97          \\
		& ResNet34 & 94.44         & 92.15         & 93.28                             & 93.31                                       & -                                    & \textbf{93.67} \\ \hline
		\multirow{2}{*}{ImageNet} & ResNet18 & 69.40         & 61.63         & 68.01                             & 69.48                                       & 68.99                                & \textbf{69.02} \\
		& ResNet50 & 76.48         & 68.94         & 74.73                             & 74.45                                       & \textbf{75.51}                       & 74.84          \\ \bottomrule
	\end{tabular}
	\label{resnet}
\end{table}
\begin{table}[]
	\centering
	\scriptsize
	\renewcommand{\arraystretch}{1.0}
	\setlength{\tabcolsep}{3pt}  
	\caption{Results on Transformers. }
	\begin{threeparttable}
		\begin{tabular}{llllllllll}
			\toprule
			Dataset                  & Model                                                        & \begin{tabular}[c]{@{}l@{}}Traning \\ type\end{tabular} & Metric   & \multicolumn{2}{l}{Baselines} & \cite{xi2023training} & Ultra-low & \cite{ghaffari2022integer} & Ours           \\
			&                                                              &                                                         &          &               &               &                                        & \cite{sun2020ultra}           &                                             &                \\ \cline{5-10} 
			&                                                              &                                                         &          & FP            & INT8          & INT 4/4/4                              & Radix-4 4/4/4                                  & INT 8/8/8                                   & INT 6/6/6      \\ \hline
			\multirow{2}{*}{CIFAR10} & ViT-B                                                        & Fine-tune                                               & Top1 Acc & 98.85         & 94.03         & 98.36                                  & -                                              & -                                           & \textbf{98.40} \\
			& ViT-L                                                        & Fine-tune                                               & Top1 Acc & 98.90         & 93.69         & 98.47                                  & -                                              & 98.80                                       & \textbf{98.84} \\
			WMT14                    & \begin{tabular}[c]{@{}l@{}}Transformer-\\ based\end{tabular} & Pretrain                                                & BLEU     & 27.5          & 21.1          & \textbf{27.17}                         & 25.4                                           & -                                           & 27.09          \\ \bottomrule
		\end{tabular}
		\begin{tablenotes}
			\item[*] \cite{xi2023training} applies 8-bit integer format  on gradients and pruns half of gradients. 
		\end{tablenotes}
	\end{threeparttable}
	\label{transformer}
\end{table}

\textbf{Machine translation}
We train a transformer-based architecture\footnote{\href{https://github.com/facebookresearch/fairseq}{https://github.com/facebookresearch/fairseq}} on the WMT14 English-German (en-de) dataset \cite{bojar2014findings}. Within the entire network, the linear layers utilize 6-bit weights, activations, and gradients, along with an 8-bit fully-quantized L1 Batch Normalization (L1BN) layer. The computation of soft-max in attention mechanism is in floating-point. The results is reported in Table \ref{transformer}. Our method achieves a negligible loss of accuracy. 

\textbf{Transductive and inductive prediction}

We train a Temporal Graph Network (TGN \cite{kumar2019predicting}) on the Wikipedia \footnote{\href{http://snap.stanford.edu/jodie/wikipedia.csv}{http://snap.stanford.edu/jodie/wikipedia.csv}} dataset. All the linear layers in TGN utilize 8-bit weights, activations, and gradients. Our method even surpasses the performance on full-precision training. 

\textbf{Time series prediction}

We train a Recurrent Neural Network (RNN \cite{medsker2001recurrent}) on the ElectricityLoadDiagrams2011-2014 \footnote{\href{https://archive.ics.uci.edu/dataset/321/electricityloaddiagrams20112014}{https://archive.ics.uci.edu/dataset/321/electricityloaddiagrams20112014}} dataset for time series prediction task. Our method achieves a negligible loss of accuracy. 

Our method outperforms competitors across nearly all networks and tasks. In convolutional neural networks, our method achieves the smallest bitwidth. In transformers, although our method has a larger bitwidth compared to \cite{xi2023training}, it demonstrates superior hardware efficiency, which we will discuss in Sec. \ref{hardware}. We are the first to evaluate integer training on both TGN and RNN architectures. Our method demonstrates strong accuracy retention, showcasing its applicability and robustness across diverse network architectures. 

\begin{figure}[]
	\centering
	\begin{minipage}{0.4\textwidth}
		\centering
		\scriptsize
		\renewcommand{\arraystretch}{1}
		\setlength{\tabcolsep}{1.4pt}
			\captionof{table}{Results on Temporal Graph Network (TGN \cite{kumar2019predicting}) . }
		\begin{tabular}{lllll}
			\toprule
			Dataset                    & Metric & Full-Precision & StateQuant & Ours  \\ \hline
			\multirow{2}{*}{Wikipedia} & AUC    & 0.936          & 0.939      & 0.939 \\
			& AP     & 0.945          & 0.945      & 0.947 \\ \bottomrule
		\end{tabular}
		\label{gnn}
	\end{minipage}
	\hfill
	\begin{minipage}{0.58\textwidth}
		\centering
		\scriptsize
		\renewcommand{\arraystretch}{1}
		\setlength{\tabcolsep}{1.4pt}
		\captionof{table}{Results on Recurrent Neural Network (RNN \cite{medsker2001recurrent}) . }
		\begin{tabular}{lllll}
			\toprule
			Dataset                                          & Metric & Full-Precision & StateQuant & Ours  \\ \hline
			\multirow{2}{*}{ElectricityLoadDiagrams20112014} & ND     & 0.069          & 0.078      & 0.074 \\
			& RMSE   & 0.526          & 0.516      & 0.475 \\ \bottomrule
		\end{tabular}
		\label{rnn}
	\end{minipage}
\end{figure}

\subsection{Parameter Analysis and Ablation study}
\begin{figure}[]
	\centering
	\begin{minipage}{0.58\textwidth}
		\centering
		\scriptsize
		\renewcommand{\arraystretch}{1}
		\setlength{\tabcolsep}{1.4pt}
		\captionof{table}{Performance under different number of groups. }
		\begin{tabular}{lllllll}
			\toprule
			number of groups & 6     & 5     & 4     & 3 & 2 & 1 \\ \hline
			ResNet20         & 90.47 & 90.47 & 90.41 & 89.71  & 88.39  & 86.45  \\
			ViT-B            & 98.57 & 98.52 & 98.40 & 98.36  & 97.54  & 96.65  \\ \bottomrule
		\end{tabular}
		\label{groups}
	\end{minipage}
	\hfill
	\begin{minipage}{0.4\textwidth}
		\centering
		\scriptsize
		\renewcommand{\arraystretch}{1}
		\setlength{\tabcolsep}{1.4pt}
		\captionof{table}{Ablation on L1 normalization. }
		\begin{tabular}{llllll}
			\toprule
			& FP    & L2BN  & L1BN  & QL2BN & QL1BN \\ \hline
			ResNet20 & 91.25 & 90.78 & 90.95 & 13.34 & 90.40 \\
			ResNet38 & 92.04 & 91.43 & 91.67 & 11.92 & 91.46 \\ \bottomrule
		\end{tabular}
		\label{ablation_l1}
	\end{minipage}
\end{figure}
\textbf{Analysis on number of groups}

In ShiftQuant, an increase in the number of groups leads to finer quantization granularity, but an excessive number of groups can also increase the computational burden. We conducte experiments on ResNet20 and ViT-B using different group settings on the CIFAR10 dataset. As shown in Table \ref{groups}, the performance improvement reduces when number of groups increases. When the number of groups is 4, the grouping map can be represented using exactly 2 bits, while maintaining relatively high precision. Therefore, we set the number of groups as 4.

\textbf{Ablation study on L1 normalization}

As shown in Table \ref{ablation_l1}, fully-quantized L2 normalization layers are not suitable for  low-precision  training. As a contrast, our fully-quantized L1 normalization layers achieve competitive performance. 
\subsection{Hardware Overhead}\label{hardware}

\textbf{Throughput analysis on ARMv8}

\begin{figure}[h]
	\centering 
	\includegraphics[width=1\linewidth, height=0.3\linewidth]{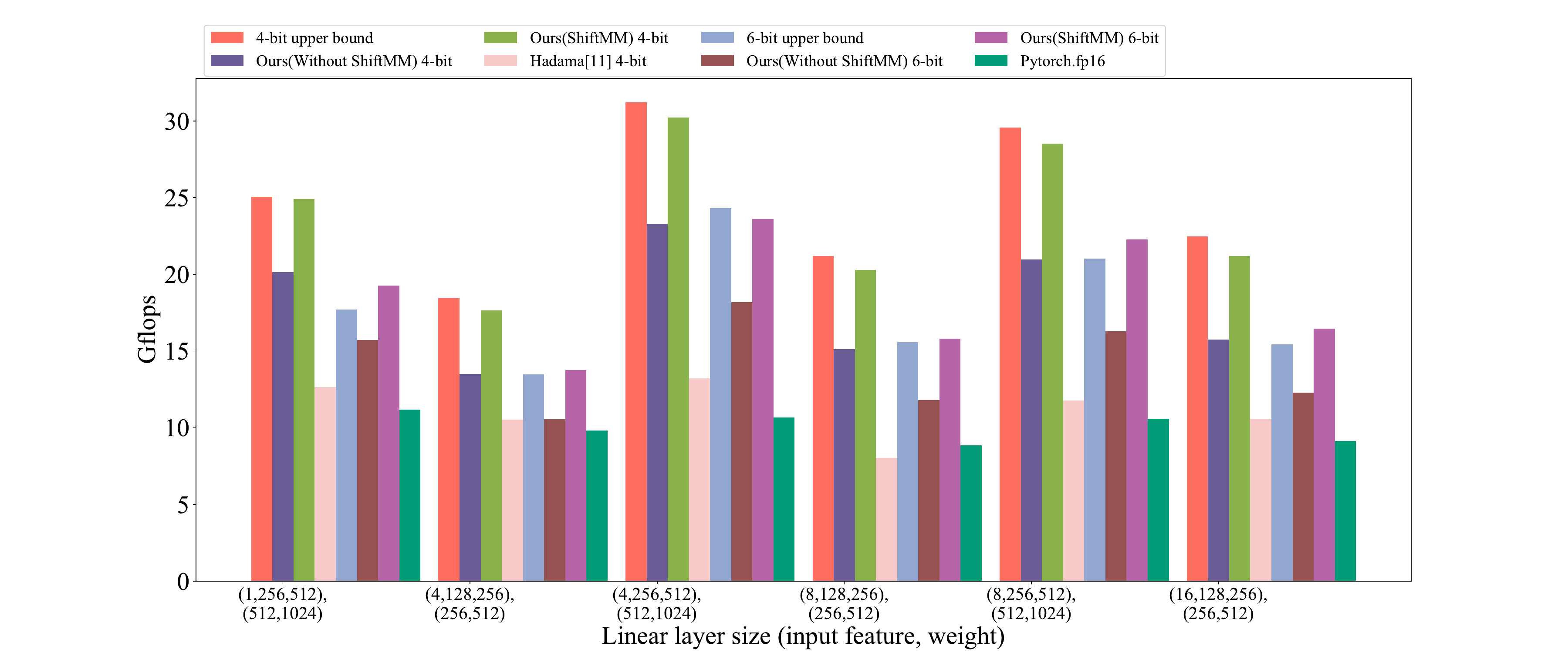}
	\caption{Comparison of simplest per-tensor quantization (the upper bound of efficiency), ShiftQuant (GEMM implementation and ShiftMM implementation), \cite{xi2023training}, Pytorch.fp16 on ARMv8. }
	\label{cpu}
\end{figure}

To evaluate the efficiency of our method, we construct two baselines. The first is a fully-quantized linear layer, which utilizes the simplest per-tensor quantization and implements matrix multiplication by widely-used OpenBLAS\footnote{\hyperlink{https://www.openblas.net/}{https://www.openblas.net/}}. This implementation reflects the upper bounds of efficiency under a given bitwidth. The second is made against a model using torch.fp16 precision. Meanwhile, we implement one comparative method \cite{xi2023training}.  We assess the throughput across various sizes of linear layers. As depicted in Figure \ref{cpu}, both the GEMM and ShiftMM implementations of our method significantly outperform the torch.fp16 model. ShiftMM closely matches the performance of the baseline 4-bit implementation. The performance gap between GEMM and ShiftMM implementation comes from memory rearrangement. Moreover, 6-bit ShiftMM outperforms 4-bit \cite{xi2023training}. Two ingredients lead to this, the extra computation before quantization and hardware-unfriendly pruning in \cite{xi2023training}. 

\textbf{Throughput analysis on GPU}

\begin{figure}[h]
	\centering
	\begin{minipage}{0.48\textwidth}
		\includegraphics[width=1\linewidth, height=0.45\linewidth]{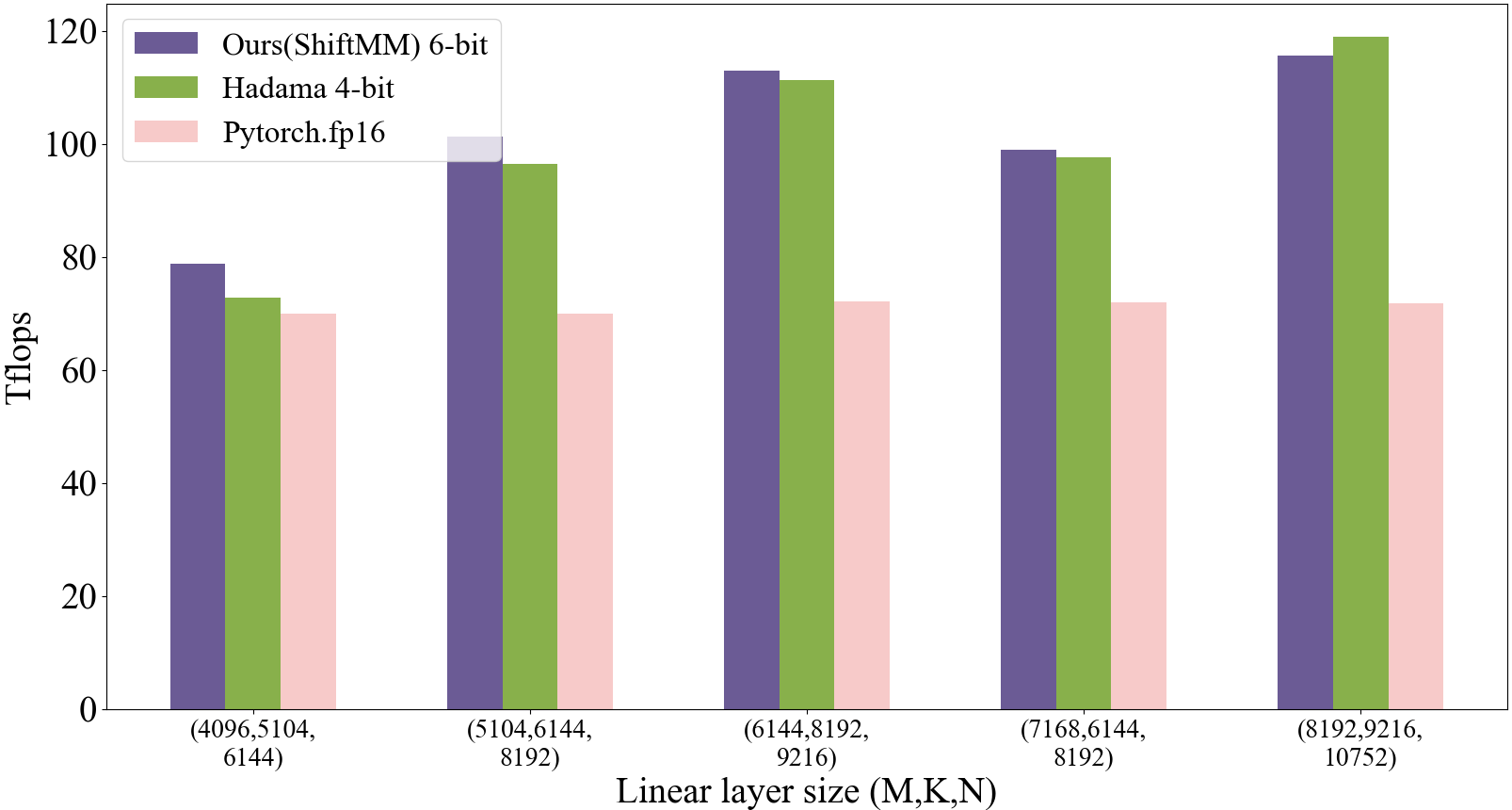}
		\caption{Throughput of ShiftMM, Hadama \cite{xi2023training}, and Pytorch.fp16 on Nvidia RTX 3090. }
		\label{gpu}
	\end{minipage}
	\hfill
	\begin{minipage}{0.51\textwidth}
		\includegraphics[width=1\linewidth, height=0.45\linewidth]{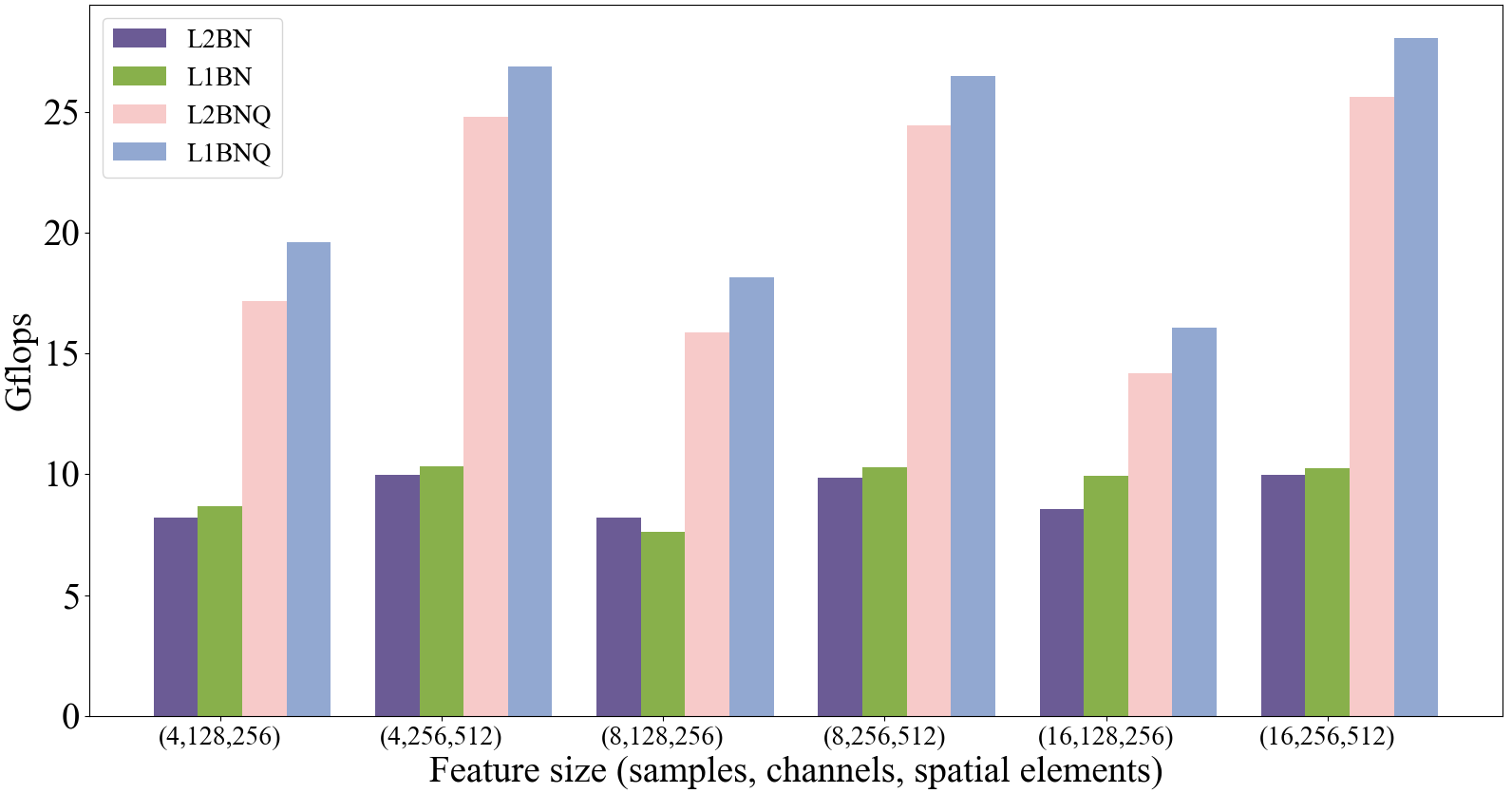}
		\caption{Comparision of different normalization layers on ARMv8. }
		\label{bn_cpu}
	\end{minipage}
\end{figure}

As shown in Figure \ref{gpu}, we analyze the performance of our 6-bit ShiftMM, 4-bit \cite{xi2023training}, and Pytorch.fp16 on Nvidia RTX 3090. As the size grows, ShiftMM achieves more performance improvement than pytorch.fp16. Due to the heavy reflection operation and pruning, \cite{xi2023training} does not take full advantage of the low bitwidth.  

\textbf{Resource consumption on FPGA}

To validate the efficiency of ShiftMM, we analyze the resource consumption of ShiftMM and basic GEMM on Xilinx ZC706 board. We perform a matrix multiplication with size (1024, 288, 32) on FPGA. For a comprehensive comparison, we select two resource bind strategy, LUT priority and DSP priority. We apply DSP reusing technique on implementation (details can be found in Appendix. \ref{dspreusing}). As shown in Table \ref{fpga}, in LUT priority setting, our INT6 implementation saves 60.7$\%$ FF and 43.5$\%$ LUT compared with FP16 implementation. In DSP reusing setting, we utilize DSP reusing technique. Our INT6 implementation saves 50$\%$ DSP, 55.3$\%$ FF with $3.4\%$ overhead on LUT compared with FP16 implementation. Meanwhile, ShiftMM achieves closed resource utilization rate with the basic GEMM. 
\begin{table}[h]
	\centering
	\small
	\setlength{\tabcolsep}{3.5pt}
	\caption{Resource consumption of implementation on FPGA (ZC706). }
\begin{tabular}{l|llllll|llllll}
	\toprule
	& \multicolumn{6}{l|}{LUT priority}                                                                                                  & \multicolumn{6}{l}{DSP priority}                                                                                                  \\ \hline
	Method          & \multicolumn{4}{l|}{Baseline}                                                                          & \multicolumn{2}{l|}{Ours} & \multicolumn{4}{l|}{Baseline}                                                                          & \multicolumn{2}{l}{Ours} \\ \hline
	Bitwidth        & \multicolumn{1}{l|}{4} & \multicolumn{1}{l|}{6} & \multicolumn{1}{l|}{8} & \multicolumn{1}{l|}{16(FP)} & 4           & 6           & \multicolumn{1}{l|}{4} & \multicolumn{1}{l|}{6} & \multicolumn{1}{l|}{8} & \multicolumn{1}{l|}{16(FP)} & 4           & 6          \\ \hline
	FF              & 414                    & 576                    & 744                    & \multicolumn{1}{l|}{1504}   & 414         & 590         & 818                    & 1042                   & 1362                   & \multicolumn{1}{l|}{2978}   & 994         & 1330       \\ \hline
	DSP             & 0                      & 0                      & 0                      & \multicolumn{1}{l|}{0}      & 0           & 0           & 8                      & 16                     & 16                     & \multicolumn{1}{l|}{32}     & 8           & 16         \\ \hline
	LUT             & 4671                   & 5050                   & 5406                   & \multicolumn{1}{l|}{9178}   & 4799        & 5190        & 4595                   & 4723                   & 5619                   & \multicolumn{1}{l|}{5403}   & 5203        & 5587       \\ \hline
	Latency (ms) & 5.95                 & 6.32                 & 6.32                 & \multicolumn{1}{l|}{6.32} & 5.95      & 5.95      & 4.63                 & 4.63                 & 4.63                 & \multicolumn{1}{l|}{4.82} & 4.63      & 4.63     \\ \bottomrule
\end{tabular}
	\label{fpga}
\end{table}

\textbf{Time proportion for each part of ShiftQuant}

\begin{figure}[h]
	\centering 
	\includegraphics[width=0.5\linewidth, height=0.3\linewidth]{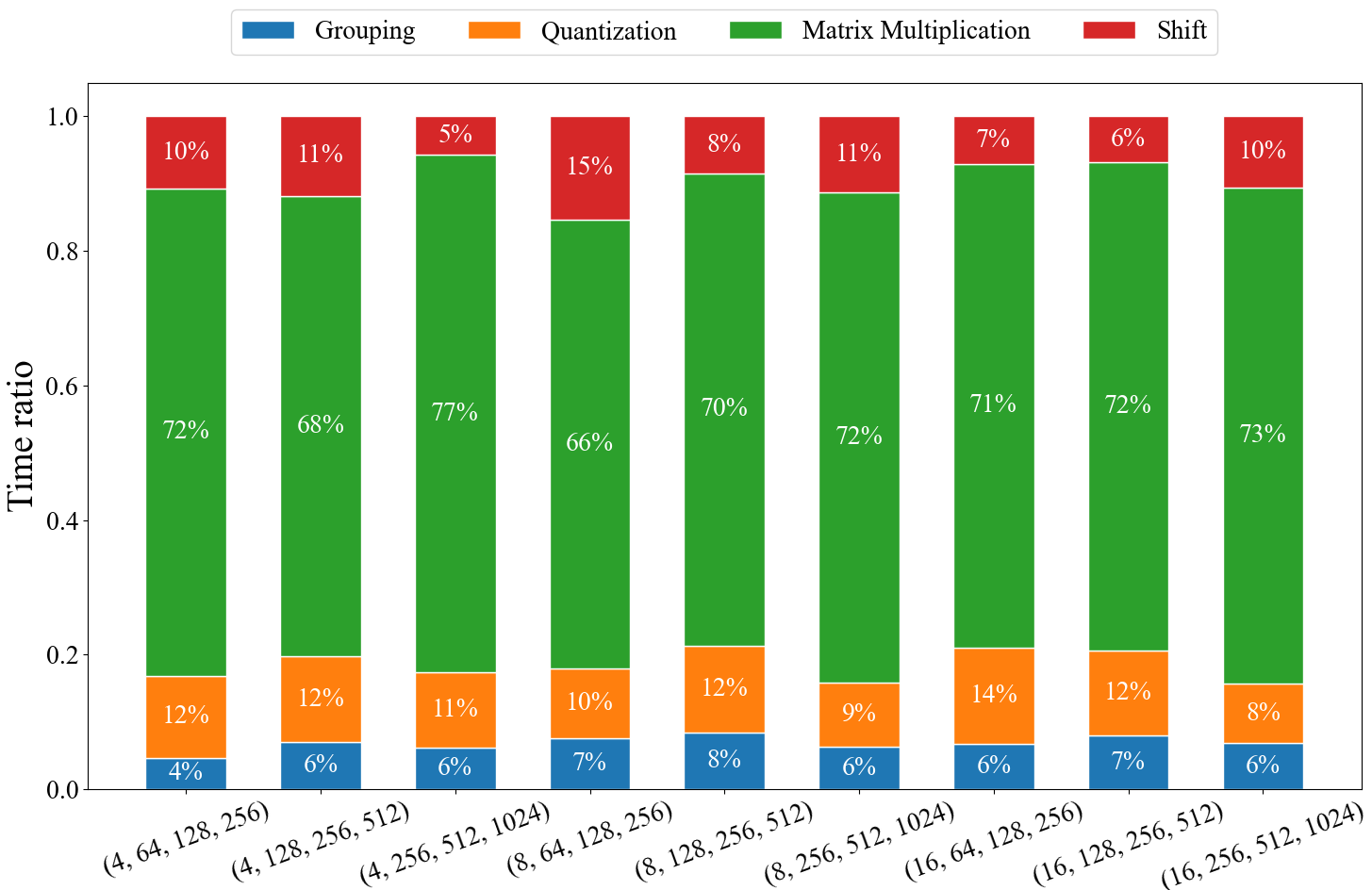}
	\caption{Time proportion for each part of ShiftQuant. }
	\label{percent}
\end{figure}

We evaluate the time proportion for each part of ShiftQuant on ARMV8. As depicted in Figure \ref{percent}, the power-of-two grouping strategy and shift operation contribute minimally to the overall latency, with the majority of the time consumed by matrix computations. This observation aligns with the criteria for an efficient quantizer and further validates the superiority of ShiftQuant. 

\textbf{End-to-end acceleration performance}

As shown in Table \ref{endtoend}, we evaluate the end-to-end training acceleration of the Vision Transformer (ViT) across various batch sizes on ARMV8 and RTX-3090. The subsequent tables illustrate the acceleration achieved by our method compared to FP16 counterparts. Our method markedly accelerates training. It is important to note that this was achieved through a basic implementation, without fully leveraging the potential of ShiftQuant. With further engineering optimizations, the acceleration performance could be significantly enhanced. 

\begin{table}[h]
	\centering
	\normalsize
	\setlength{\tabcolsep}{3.5pt}
	\caption{End-to-end acceleration on different platforms. }
	\begin{tabular}{l|llll|lllcllll}
		\toprule
		Platform  & \multicolumn{4}{l|}{ARMV8} & \multicolumn{8}{l}{RTX3090}                                                                                                          \\ \hline
		Batchsize & 1    & 4     & 8    & 16   & 32                      & 40                       & 48                       & 64                       & 80   & 96   & 128  & 192  \\ \hline
		ViT-B-16  & 2\%  & 14\%  & 26\% & 35\% & \multicolumn{1}{c}{9\%} & \multicolumn{1}{c}{12\%} & \multicolumn{1}{c}{15\%} & 21\%                     & 18\% & 26\% & 34\% & 62\% \\
		ViT-L-16  & 3\%  & 18\%  & 31\% & 43\% & 12\%                    & 19\%                     & 23\%                     & \multicolumn{1}{l}{26\%} & 28\% & 33\% & 39\% & 68\% \\ \bottomrule
	\end{tabular}
\label{endtoend}
\end{table}

\section{Conclusion}
In this paper, we enhance sub-8-bit integer training from two aspects. We propose ShiftQuant to eliminate quantization noise in gradient estimation, and introduce fully-quantized L1 normalization layers to smoothen the loss landscape for stable convergence. Comprehensive experiments validate the efficiency and accuracy of our method. Meanwhile, we have implemented ShiftQuant on multiple types of devices and proven its applicability. The first future direction is to further improve accuracy and efficiency, including developments in algorithms and implementation techniques. The second future direction is to apply ShiftQuant to other tasks, such as inference acceleration of large language models (LLMs). The excellent outlier suppression capacity of ShiftQuant may be useful for other challenging quantization tasks. 


\bibliographystyle{iclr2025_conference}

\newpage
\appendix
\section*{Appendix}

\section{Proof of the Equivalence between Problem (1) and the Direct Optimization of Quantization Variance}\label{equal}

Following Eq. \ref{sq}, given a variable $x$, the quantization variance under stochastic rounding is: 
\begin{equation}\label{varianceq}
	\begin{aligned}
		Var[Q(x)] &= (l(x)-x)^2\cdot p_Q(x) + (u(x)-x)^2\cdot (1-p_Q(x)) \\
		&= (x-l(x)) \cdot (u(x)-x), 
	\end{aligned}
\end{equation}
where $l(x)$ and $u(x)$ are lower quantization level and upper quantization level of $x$. 

From the viewpoint of possibility, minimizing quantization variance is equal to minimize the expectation of quantization variance. The object function is: 
\begin{equation}
	\min\,E[Var(Q(x))]. 
\end{equation}
For $Q_N$ levels quantization, above function can be represented as: 
\begin{equation}\label{evar}
	\begin{aligned}
		E[Var(Q(x))] & = \sum_{m=1}^{Q_N} \{ \int_{l_m}^{u_m} (x-l_m) \cdot (u_m-x)\} \\
		& = \sum_{m=1}^{Q_N} \{ \int_{l_m}^{u_m} (l_m+u_m)\cot x \cdot p(x)dx- \int_{l_m}^{u_m} x^2\cdot p(x)dx  - \int_{l_m}^{u_m} l_m u_m \cdot p(x)dx\}. 
	\end{aligned}
\end{equation}
$p(x)$ is the distribution of $x$. Any distribution can be representative as the mix of Laplace distributions with different parameters. Thus, $p(x)$ can be represent as:
\begin{equation}\label{mlaplace}
	\begin{aligned}
		p(x) =\sum_{k=1}^{\inf} \alpha_k\frac{1}{2\lambda_k}e^{-\frac{|x-\mu_k|}{\lambda_k}}. 
	\end{aligned}
\end{equation}
Insert Eq. \ref{mlaplace} to Eq. \ref{evar}:
\begin{equation}
	\begin{aligned}
		E[Var(Q(x))] & = \sum_{k=1}^{\inf}\alpha_k\sum_{m=1}^{Q_N} \{ \int_{l_m}^{u_m} (l_m+u_m)\cot x \cdot \frac{1}{2\lambda_k}e^{-\frac{|x-\mu_k|}{\lambda_k}}dx \\ & - \int_{l_m}^{u_m} x^2\cdot \frac{1}{2\lambda_k}e^{-\frac{|x-\mu_k|}{\lambda_k}}dx  \\ & - \int_{l_m}^{u_m} l_m u_m \cdot \frac{1}{2\lambda_k}e^{-\frac{|x-\mu_k|}{\lambda_k}}dx\}.  
	\end{aligned}
\end{equation}
We first analyze on one Laplace component:
\begin{equation}
	\begin{aligned}
		t_k(x) & = \sum_{m=1}^{Q_N} \{ \int_{l_m}^{u_m} (l_m+u_m)\cot x \cdot \frac{1}{2\lambda_k}e^{-\frac{|x-\mu_k|}{\lambda_k}}dx \\ & - \int_{l_m}^{u_m} x^2\cdot \frac{1}{2\lambda_k}e^{-\frac{|x-\mu_k|}{\lambda_k}}dx  \\ & - \int_{l_m}^{u_m} l_m u_m \cdot \frac{1}{2\lambda_k}e^{-\frac{|x-\mu_k|}{\lambda_k}}dx\}, let z= \frac{x-\mu_k}{\lambda_k} \\
		& = \sum_{m=1}^{Q_N} \overbrace{\{\frac{l_m+u_m}{2}\int_{\frac{l_m-\mu_k}{\lambda_k}}^{\frac{u_m-\mu_k}{\lambda_k}}(\lambda_k z+\mu_k)e^{-|z|}dz}^{f^m(x)} \\
		& - \overbrace{\frac{1}{2}\int_{\frac{l_m-\mu_k}{\lambda_k}}^{\frac{u_m-\mu_k}{\lambda_k}}(\lambda_k z+\mu_k)^2e^{-|z|}dz}^{g^m(x)} \\
		& - \overbrace{\frac{l_m u_m}{2}\int_{\frac{l_m-\mu_k}{\lambda_k}}^{\frac{u_m-\mu_k}{\lambda_k}}e^{-|z|}dz}^{q^m(x)}
		\}. 
	\end{aligned}
\end{equation}
\textbf{Analysis on $f^m(x)$}

\begin{equation}
	\int_{\frac{l_m-\mu_k}{\lambda_k}}^{\frac{u_m-\mu_k}{\lambda_k}}(\lambda_k z+\mu_k)e^{-|z|}dz=
	\begin{cases} 
		\lambda_k[(-\frac{u_m-\mu_k}{\lambda_k}+1)e^{-\frac{u_m-\mu_k}{\lambda_k}}-(-l+1)e^{-\frac{l_m-\mu_k}{\lambda_k}}]&\\
		+\mu_k[e^{-\frac{l_m-\mu_k}{\lambda_k}}-e^{-\frac{u_m-\mu_k}{\lambda_k}}] &, \frac{u_m-\mu_k}{\lambda_k} > \frac{l_m-\mu_k}{\lambda_k} >0 \\ 
		2(\mu_k-\lambda_k)-(\lambda_k\frac{l_m-\mu_k}{\lambda_k}+\mu_k-\lambda_k)e^{\frac{l_m-\mu_k}{\lambda_k}}&\\-(\lambda_k\frac{u_m-\mu_k}{\lambda_k}+\mu_k-\lambda_k)e^{-\frac{u_m-\mu_k}{\lambda_k}} &, \frac{u_m-\mu_k}{\lambda_k}>0>\frac{l_m-\mu_k}{\lambda_k}  \\
		\lambda_k[(\frac{u_m-\mu_k}{\lambda_k}-1)e^{\frac{u_m-\mu_k}{\lambda_k}}-(l-1)e^{\frac{l_m-\mu_k}{\lambda_k}}]&\\
		+\mu_k[e^{\frac{u_m-\mu_k}{\lambda_k}}-e^{\frac{l_m-\mu_k}{\lambda_k}}]	&, \frac{l_m-\mu_k}{\lambda_k} < \frac{u_m-\mu_k}{\lambda_k} < 0. 
	\end{cases}
\end{equation}
With assumption of symmetry on quantization range, we have: 
\begin{equation}
	\begin{aligned}
		\sum_{m=1}^{Q_N}\int_{\frac{l_m-\mu_k}{\lambda_k}}^{\frac{u_m-\mu_k}{\lambda_k}}(\lambda_k z+\mu_k)e^{-|z|}dz= & 
		\lambda_k[(1-\frac{u_{Q_N}-\mu_k}{\lambda_k})e^{\frac{u_{Q_N}-\mu_k}{\lambda_k}}-(l_1-1)e^{\frac{l_1-\mu_k}{\lambda_k}}] \\
		&+\mu_k[e^{\frac{l_1-\mu_k}{\lambda_k}}-e^{-\frac{u_{Q_N}-\mu_k}{\lambda_k}}]+2(\mu_k-\lambda_k). 
	\end{aligned}
\end{equation}
For clarity, we represent above function as $\beta(u_{Q_N},l_1),\beta(u_{Q_N},l_1)=\sum_{m=1}^{Q_N}\int_{\frac{l_m-\mu_k}{\lambda_k}}^{\frac{u_m-\mu_k}{\lambda_k}}(\lambda_k z+\mu_k)e^{-|z|}dz$. Then, we have: 
\begin{equation}\label{fmx}
	\begin{aligned}
		\sum_{m=1}^{Q_N}f^m(x) & = (\tau-\frac{\tau}{Q_N})\beta(u_{Q_N}, l_0)-\frac{2\tau}{Q_N}\sum_{m=1}^{Q_N}f(u_m,l_1) \\
		&=\frac{Q_N\tau-\tau}{Q_N}\cdot2\lambda\cdot e^{-\frac{u_{Q_N}-\mu_k}{\lambda_k}}-\frac{2\tau}{Q_N}\sum_{m=1}^{Q_N}\lambda[e^{-|u_m|}  \\&+e^{-\frac{l_{1}-\mu_k}{\lambda_k}}-|u_m|e^{-|u_m|}-\frac{l_{1}-\mu_k}{\lambda_k}e^{-\frac{l_{1}-\mu_k}{\lambda_k}}-2] \\
		&=2\lambda\tau\frac{Q_N-1}{Q_N}e^{-\frac{u_{Q_N}-\mu_k}{\lambda_k}}-2\lambda\tau\frac{1}{Q_N}\cdot(Q_N-1)\cdot[e^{\frac{l_{1}-\mu_k}{\lambda_k}}-\\ &\frac{l_{1}-\mu_k}{\lambda_k}e^{-\frac{l_{1}-\mu_k}{\lambda_k}}-2] - \frac{2\lambda\tau}{Q_N}\sum_{m=1}^{Q_N}[e^{-|u_m|}-|u_m|e^{-|u_m|}]\\
		&=2\lambda\tau\frac{Q_N-1}{Q_N}(\frac{u_{Q_N}-\mu_k}{\lambda_k}e^{-\frac{u_{Q_N}-\mu_k}{\lambda_k}})- \frac{2\lambda\tau}{Q_N}\sum_{m=1}^{Q_N}[e^{-|u_m|}-|u_m|e^{-|u_m|}], \\
		&\approx2\lambda\tau\frac{Q_N-1}{Q_N}(\frac{u_{Q_N}-\mu_k}{\lambda_k}e^{-\frac{u_{Q_N}-\mu_k}{\lambda_k}})
	\end{aligned}
\end{equation}
where $\tau$ is the quantization range, and $u_{Q_N}=\frac{\tau}{2}, l_1=-\frac{\tau}{2}$. 

\textbf{Analysis on $g^m(x)$}

We first analyze $\int_{\frac{l_m-\mu_k}{\lambda_k}}^{\frac{u_m-\mu_k}{\lambda_k}}(\lambda_k z+\mu_k)^2e^{-|z|}dz$. Let $l=\frac{l_m-\mu_k}{\lambda_k}, u=\frac{u_m-\mu_k}{\lambda_k}$, we can find: 
\begin{equation}
	\int_{l}^{u}(\lambda_k z+\mu_k)^2e^{-|z|}dz= \\
	\begin{cases} 
		\lambda^2[-z^2e^{-z}-2ze^{-z}-2e^{-z}]|^u_l+\\2\lambda\mu[-ze^{-z}-e^{-z}]|^u_l+\mu^2[-e^{-z}]|^u_l, & u>l>0 \\
		\lambda^2[(z^2-2z+2)e^z|^0_l+(-z^2+2z-2)e^{-z}|^u_0]+\\2\lambda\mu[(z-1)e^z|^0_l+(-z+1)e^{-z}|^u_0]+\mu^2[e^z|^0_l+(-e^{-z})|^u_0], & u>0>l \\
		\lambda^2[z^2e^z-2ze^z+2e^z]|^u_l+\\2\lambda\mu[ze^z-e^z]|^u_l+\mu^2[e^z]|^u_l, & l<u<0. 
	\end{cases}
\end{equation}
Accumulating the integers around different quantization levels: 
\begin{equation}\label{gmx}
	\begin{aligned}
		\sum_{m=1}^{Q_N}\int_{\frac{l_m-\mu_k}{\lambda_k}}^{\frac{u_m-\mu_k}{\lambda_k}}(\lambda_k z+\mu_k)^2e^{-|z|}dz & =\lambda^2[-e^{-z}(z^2+2z+2)]|^u_0+\mu^2[-e^{-z}]|^u_0+\lambda^2[e^z(z^2-2z+2)]|^0_l+\mu^2[e^z]|^0_l \\
		&=\lambda^2[-e^{-\frac{u_m-\mu_k}{\lambda_k}}(\frac{u_m-\mu_k}{\lambda_k}^2+2\frac{u_m-\mu_k}{\lambda_k}+2)] \\
		&+\mu^2[-e^{-\frac{u_m-\mu_k}{\lambda_k}}+1]+\lambda^2[2-e^{\frac{l_m-\mu_k}{\lambda_k}}(\frac{l_m-\mu_k}{\lambda_k}^2-2\frac{l_m-\mu_k}{\lambda_k}+2)] \\
		&+\mu^2[1-e^{\frac{l_m-\mu_k}{\lambda_k}}] \\
		&=2\mu^2(-e^{-\frac{u_m-\mu_k}{\lambda_k}}+1) \\
		&+2\lambda^2[-e^{-\frac{u_m-\mu_k}{\lambda_k}}(\frac{u_m-\mu_k}{\lambda_k}+1)^2-e^{-\frac{u_m-\mu_k}{\lambda_k}}+2]. 
	\end{aligned}
\end{equation}

\textbf{Analysis on $q^m(x)$}

We first analyze $\int_{\frac{l_m-\mu_k}{\lambda_k}}^{\frac{u_m-\mu_k}{\lambda_k}}e^{-|z|}dz$. Let $l=\frac{l_m-\mu_k}{\lambda_k}, u=\frac{u_m-\mu_k}{\lambda_k}$, we can find: 
\begin{equation}
	\int_{l}^{u}e^{-|z|}dz = \\
	\begin{cases}
		e^{-l}-e^{-u}, &u>l>0 \\
		2-e^{l}-e^{-u}, &u>0>l \\
		e^u-e^l, &l<u<0. 
	\end{cases}
\end{equation}
Thus we have:
\begin{equation}
	\sum_{m=1}^{Q_N}\int_{\frac{l_m-\mu_k}{\lambda_k}}^{\frac{u_m-\mu_k}{\lambda_k}}e^{-|z|} = 2-2e^{-u}
\end{equation}
Following the same derivation as Eq. (\ref{fmx}), we can find:
\begin{equation}\label{qmx}
	\sum_{m=1}^{Q_N}\frac{l_m+u_m}{2}\int_{\frac{l_m-\mu_k}{\lambda_k}}^{\frac{u_m-\mu_k}{\lambda_k}}e^{-|z|} = \frac{\tau\cdot (\tau-\frac{2\tau}{Q_N})}{2}(2-2e^{-u}). 
\end{equation}

Combine the result of Eq. (\ref{fmx}), Eq. (\ref{gmx}) and Eq. (\ref{qmx}), we can find: 

\begin{equation}\label{finalv}
	\begin{aligned}
		t_k(x) &= 2\lambda\tau\frac{Q_N-1}{Q_N}(ue_{-u}-2)-\{\mu^2(-e^{-u}+1)+\\
		&\lambda^2[-e^{-u}(u+1)^2-e^{-u}+2]\}-\frac{\tau\cdot (\tau-\frac{2\tau}{Q_N})}{2}(2-2e^{-u})\\
		&= 2\lambda\tau\frac{Q_N-1}{Q_N}(ue_{-u}-2)+\mu^2(e^{-u}-1)+
		\\&\lambda^2[e^{-u}(u+1)^2+e^{-u}-2]+\frac{\tau\cdot (\tau-\frac{2\tau}{Q_N})}{2}(2e^{-u}-2),
	\end{aligned}
\end{equation}
where $u=\frac{u_{Q_N}-\mu_k}{\lambda_k}$ and $u\gg 1$. Therefore, $t_k(x)$ is a monotonically increasing function of $u$ at a single point. As the quantization range $\tau$ decreases, the value of $t_k(x)$ also decreases. It is evident that the same conclusion holds on $E[Var(Q(x))]=\sum_{k=1}^{\inf}\alpha_kt_k(x)$.

Now we consider quantizing a vector $\boldsymbol{x}\in\mathcal{NHW}$. The expectation of the maximum of $x$ is \cite{arnold2008first, david2004order}:

\begin{equation}
	E[max\,\boldsymbol{x}] \approx \sum_{k=1}^{\inf}(\mu_k+\lambda_k\ln(2NHW))
\end{equation}
The maximum of $x$ is proportionate to $\lambda_k$. $E[Var(Q(x))]$ is a monotonically increasing function of $\frac{u_{Q_N}-\mu_k}{\lambda_k}$. Therefore, $E[Var(Q(x))]$ is also a monotonically increasing function of $\frac{u_{Q_N}}{\tau_{\boldsymbol{x}}}$ (In experiment, the distribution of gradients are extremely bell curve, which means $\mu_k$ is closed to 0). $\tau_{\boldsymbol{x}}$ is the magnitude of $\boldsymbol{x}$. 

Now, we expand to grouping channels. The object of channel grouping is to minimize the quantization variance:  
\begin{equation} \label{dq_var}
	\begin{aligned}
		\min_{\tau_0, \tau_1, \cdots, \tau_{N_{G}-1}, \tau_{N_{G}}} & \sum_{g=1}^{N_{G}} \sum_{i \in P^g} E[Var(\boldsymbol{x}_i)], \\
		s.t.  \, r_{max} = \tau_0 \ge & \tau_1 \ge \cdots \ge \tau_{N_{G}-1} \ge \tau_{N_{G}} = 0 , 
	\end{aligned}
\end{equation}
where $\boldsymbol{x}_i$ is the vectorized $i$-th channel. $E[Var(x_i)]$ is depend on the quantization range of the group and the range of $x_i$. As indicated before, $E[Var(x_i)]$ is a monotonically increasing function about $\frac{u_{Q_N}}{\tau_{\boldsymbol{x}_i}}$, where $\tau_{\boldsymbol{x}_i}$ is the quantization range of $\boldsymbol{x}_i$. In group quantization, $u_{Q_N}=\tau_g$, where $\tau_g$ is the upper threshold of group $g$ in Eq. (\ref{dq_f}). Hence that problem (\ref{dq_var}) and problem (\ref{dq_f}) has the same optimal point. Meanwhile, both of them are monotonically increasing about $\sum_{i \in P^g}\frac{\tau_g}{r_i}$.

\section{Theoretical Analysis on ShiftQuant}\label{anashiftquant}
\textbf{Unbiased quantizer}

ShiftQuant utilizes stochastic rounding: 
\begin{equation}\label{sq}
	SR(x)=\left\{
	\begin{array}{rcl}
		u(x)       & w.p.     & \frac{x-l(x)}{u(x)-l(x)}\\
		l(x)       & w.p.     & 1-\frac{x-l(x)}{u(l)-l(x)}
	\end{array} \right. ,
\end{equation}
where $u(x) = s^{-1}\lceil s\cdot x \rceil$ is the upper quantization level of $x$. $l(x)=s^{-1}\lfloor s\cdot x \rfloor$ is the lower quantization level of $x$. $s$ is the quantization scale. It is clear that ShiftQuant is unbiased: 
\begin{equation}
	E[D_q(X)]=E[SR((X-ZI)\cdot S)\cdot S^{-1}+ZI]=X. 
\end{equation}
$X \in \mathbb{R}^{N \times D}$ is the quantized matrix. The second dimension of $X$ is the inner dimension. $S=diag(s_{0}, \cdots , s_{D-1})$. $s_i$ is the scale applied in the $i$-th index of the inner dimension and $s_i \in \{s_l, 2s_l, \cdots, 2^{N_G-1}s_l\}$. $s_l$ is the minimum scale. $Z=diag(z_{0}, z_{1}, \cdots , z_{D-1})$ represents the zero-point matrix. $I \in \mathbb{R}^{D\times D}$ is a identity matrix. 

\textbf{Low variance}

The up bound of ShiftQuant's variance $U_{dq}$ satisfies: 
\begin{equation}\label{var_f}
	U_{dq} \le \alpha \cdot \frac{N}{4}\sum_{j=1}^{D}s_j^{-2} + 2^{-N_G} \frac{ND}{4} s^{-2}, 
\end{equation}
where $\alpha$ is depended on the distribution of channels' range, and $1\le\alpha\le4$. $\frac{N}{4}\sum_{j=1}^{D}s_j^{-2}$ is the up bound of fine-grained quantization's variance. $\frac{ND}{4} s^{-2}$ is the up bound of coarse-grained quantization's variance. Eq. (\ref{var_f}) demonstrates that ShiftQuant achieves a closed performance to fine-grained quantizers. 

\textbf{Proof of Unbiased Quantization}

\begin{equation}
	\begin{aligned}
		E[D_q(X)] & =E[SR((X-ZI)\cdot S)\cdot S^{-1}+ZI] \\
		& =E[SR(X-ZI)]\cdot S\cdot S^{-1}+ZI \\
		& =(X-ZI)\cdot S\cdot S^{-1}+ZI \\
		& =X. 
	\end{aligned}
\end{equation}
Clearly, ShiftQuant is a unbiased quantizer. 

\textbf{Proof of Low Variance}

From Eq. (\ref{varianceq}), we can find the up bound of quantization variance: 

\begin{equation}\label{varup}
	Var[SR(x)]=(x-l(x))(u(x)-x) \le \frac{[u(x)-l(x)]^2}{4},
\end{equation} 
where $u(x), l(x)$ are the neighbouring quantization levels, and $u(x)-l(x)=\frac{r}{B}=s^{-1}$. $r$ is the quantization range, and $B$ is the number of quantization bins. 

Now we expand Eq. (\ref{varup}) to $X\in\mathcal{R}^{N\times D}$. For coarse-grained quantization, we have: 

\begin{equation}
	Var[Q_{pt}(X)]=\sum_{i=1}^{N}\sum_{j=1}^{D}\frac{r^2}{4B^2}=\frac{ND}{4}s^{-2},
\end{equation}
where $r$ is the range of $X$, and $r=max\,X-min\,X$. $s_{pt}=\frac{r}{B}$ is the quantization scale. 

For per-channel quantization, we have:

\begin{equation}
	Var[Q_{pt}(X)]=\sum_{i=1}^{N}\sum_{j=1}^{D}\frac{r_j^2}{4B^2}=\frac{N}{4}\sum_{j=1}^{D}s_j^{-2},
\end{equation}
where $r_j=max\,X_{:,j}-min\,X_{:,j}$, $s_j=\frac{r_j}{B}$. 

\begin{figure}[t]
	\centering 
	\includegraphics[width=1\linewidth, height=0.5\linewidth]{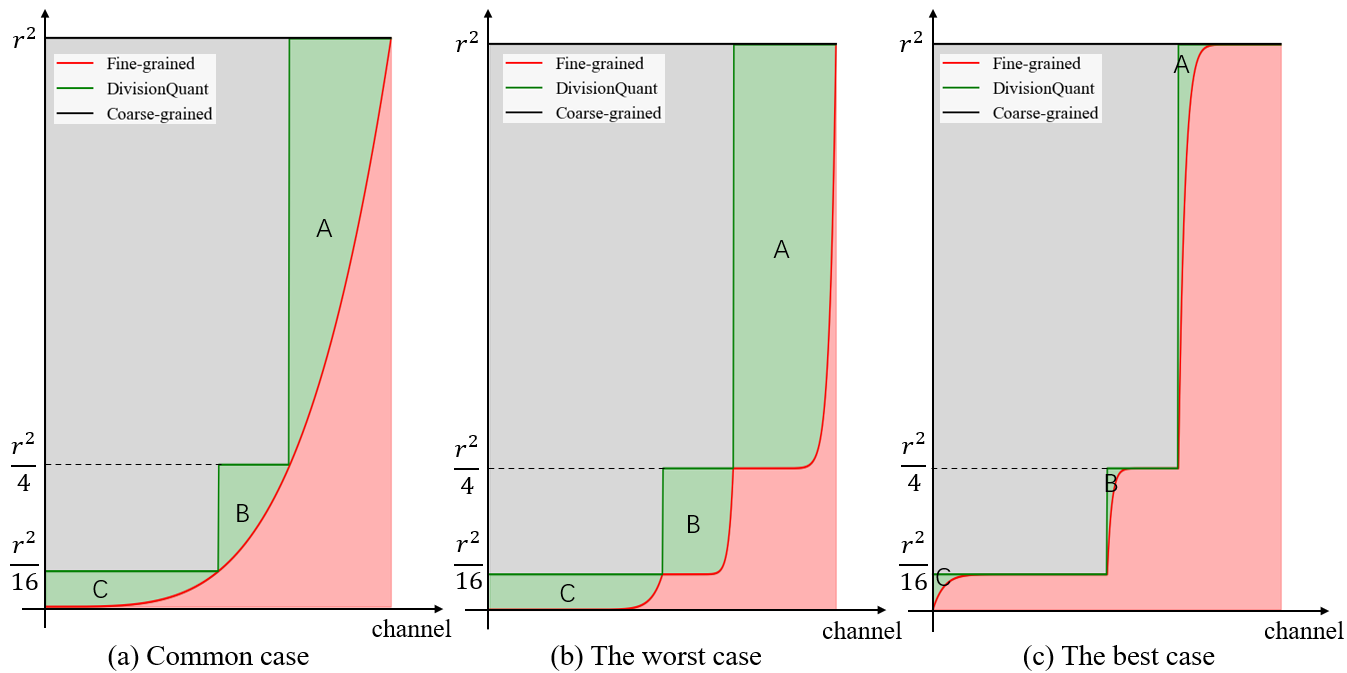}
	\caption{Quantization variance of coarse-grained quantization, fine-grained quantization, and ShiftQuant. We set number of groups as  3 in visualization. We sort the channels in ascending order by the range of channels. }
	\label{vr}
\end{figure}

We visualize the relation between fine-grained, coarse-grained quantization, and ShiftQuant in Figure \ref{vr}. 

As shown in Figure \ref{vr}(b), in the worst case, the gap of channels' ranges in a group is maximum. In this situation, we can find that: 
\begin{equation}\label{c11}
	\begin{aligned}
		U_{dq} & = U_{fq} + U_{A} + U_{B} + U_{C} \\
		& \le U_{fq} + U_{fq} * 3 + U_{C} \\
		& = U_{fq} * 4 + U_{C}, 
	\end{aligned}
\end{equation}
where $U_{fq}$ is the variance of fine-grained quantization. $U_A, U_B$, and $U_C$ denote the area of A, B, and C in Figure \ref{vr}. For $U_{C}$, we have: 
\begin{equation}\label{c12}
	U_{C} < 2^{-2*N_G+2}\cdot U_{cq},
\end{equation}
where $U_{cq}$ is the variance of coarse-grained quantization. With combining Eq. (\ref{c11}) and Eq. (\ref{c12}), we can find: 
\begin{equation}\label{c1}
	U_{dq} \le 4U_{fq} + 2^{-2*N_G+2}\cdot U_{cq}. 
\end{equation} 
As shown in Figure \ref{vr}(c), in the best case, the channel's ranges in each group is equal. The area of A, B, and C are closed to 0. The variance of ShiftQuant satisfies: 
\begin{equation}\label{c2}
	\begin{aligned}
		U_{dq} & = U_{fq} + U_{A} + U_{B} + U_{C} \\
		& \approx U_{fq} \\
		& \le U_{fq} + 2^{-2*N_G+2}\cdot U_{cq}. 
	\end{aligned}
\end{equation}
With combining Eq. (\ref{c1}) and  Eq. (\ref{c2}), we can find: 
\begin{equation}
	\begin{aligned}
		U_{dq} &\le \alpha * U_{fq} + 2^{-2*N_G+2}\cdot U_{cq}\\
		& =\alpha \cdot \frac{N}{4}\sum_{j=1}^{D}s_j^{-2} + 2^{-2N_G+2} \frac{ND}{4} s^{-2}, 
	\end{aligned}
\end{equation}
where $1\le\alpha\le4$. 

\section{Proof of L1 Normalization}

\subsection{Analysis on Lipschitzness Constant}\label{lipl1}

\begin{figure}[h]
	\centering 
	\includegraphics[width=1\linewidth, height=0.3\linewidth]{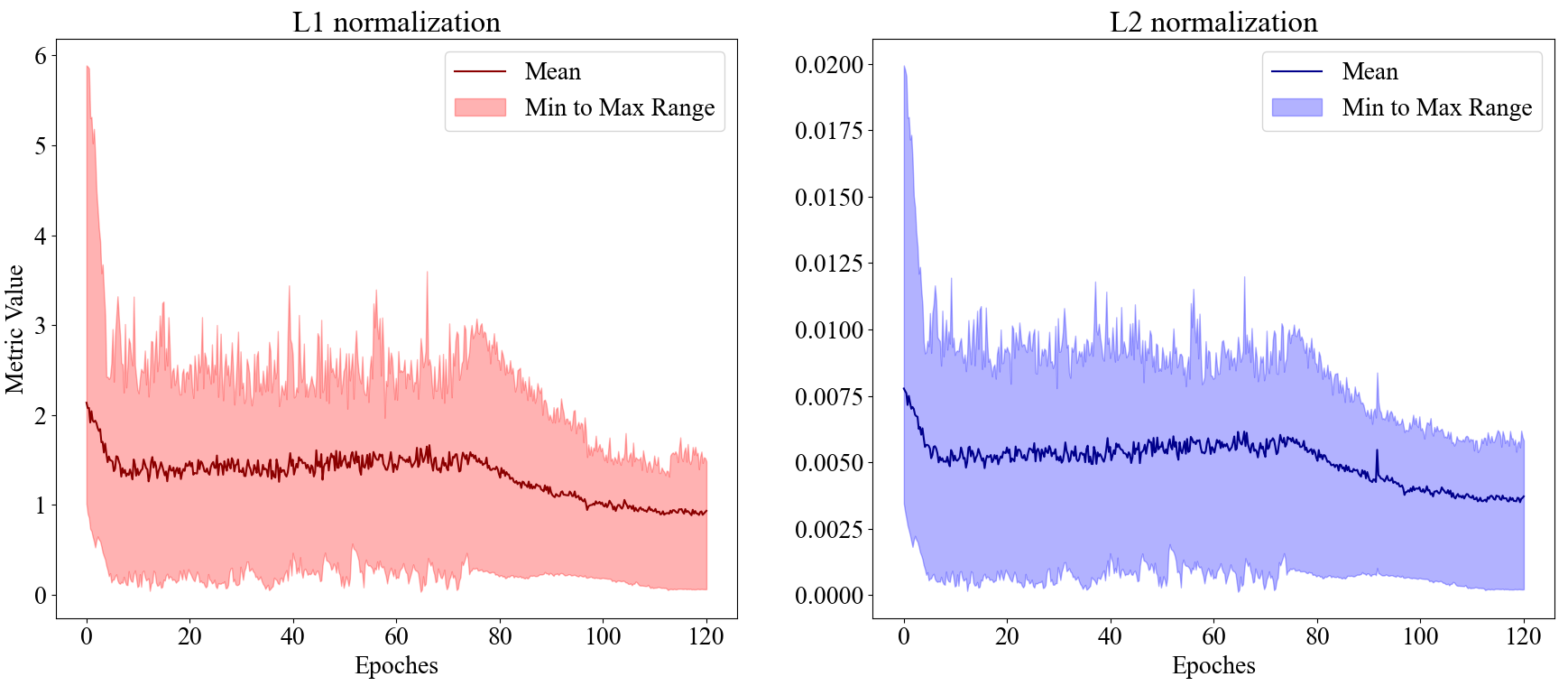}
	\caption{The L1 norm and  L2 norm of the activations during all training stage. We measure the activation before the first BN layer in ResNet20. }
	\label{l1f}
\end{figure}

A normalization layer contains two step: normalization, scale and shift:

\begin{equation}\label{normalization}
	\begin{aligned}
		\hat{\bm{x}} & = \frac{\bm{x}-\mu}{\sigma}, & Normalization,
	\end{aligned}
\end{equation}
\begin{equation}
	\begin{aligned}
		\bm{y} & = \gamma\cdot\hat{\bm{x}} + \beta, & Scale\,and\,shift, 
	\end{aligned}
\end{equation}
where $\bm{x}$ is the input vector. For example, $\bm{x}\in\mathcal{R}^{NHW}$ in batchnorm, $N,H,W$ are batchsize, height, and width of the input activation respectively. When we get $\bm{y}$, it is feed to the next layer $f(\cdot)$. Let $\bm{z}$ denote the output of the following layer ($\bm{z}=f(\bm{y})$). The gradient of $\bm{y}$ is:

\begin{equation}\label{gbn}
	\begin{aligned}
		\frac{\partial \widehat{\mathcal{L}}}{\partial \boldsymbol{x}} & =\left(\frac{\gamma}{N \sigma}\right)\left(N \frac{\partial \widehat{\mathcal{L}}}{\partial \boldsymbol{z}}-\mathbf{1}\left\langle\mathbf{1}, \frac{\partial \widehat{\mathcal{L}}}{\partial \boldsymbol{z}}\right\rangle-\hat{\boldsymbol{x}}\left\langle\frac{\partial \widehat{\mathcal{L}}}{\partial \boldsymbol{z}}, \hat{\boldsymbol{x}}\right\rangle\right) \\
		& = 
		\frac{\gamma}{\sigma}\left(\left(\frac{\partial \widehat{\mathcal{L}}}{\partial \boldsymbol{z}}-\mathbf{1} \mu_g\right)-\frac{\hat{\boldsymbol{x}}}{\left\|\hat{\boldsymbol{x}}\right\|}\left\langle\left(\frac{\partial \widehat{\mathcal{L}}}{\partial \boldsymbol{z}}-\mathbf{1} \mu_g\right), \frac{\hat{\boldsymbol{x}}}{\left\|\hat{\boldsymbol{x}}\right\|}\right\rangle\right)
		,
	\end{aligned}
\end{equation}
where $\widehat{\mathcal{L}}$ is the loss function. $\mu_g=\left\langle\mathbf{1}, \frac{\partial \widehat{\mathcal{L}}}{\partial \boldsymbol{z}}\right\rangle$. The derivation of Eq. (\ref{gbn}) can be found in \cite{santurkar2018does}. Then we have:  

\begin{equation}\label{lipg}
	\begin{aligned}
		\left\|\frac{\partial \widehat{\mathcal{L}}}{\partial \boldsymbol{x}}\right\|^2 &=\frac{\gamma^2}{\sigma^2}\left\|\left(\frac{\partial \widehat{\mathcal{L}}}{\partial \boldsymbol{z}}-\mathbf{1} \mu_g\right)-\frac{\hat{\boldsymbol{x}}}{\left\|\hat{\boldsymbol{x}}\right\|}\left\langle\left(\frac{\partial \widehat{\mathcal{L}}}{\partial \boldsymbol{z}}-\mathbf{1} \mu_g\right), \frac{\hat{\boldsymbol{x}}}{\left\|\hat{\boldsymbol{x}}\right\|}\right\rangle\right\|^2 \\
		& \le\frac{\gamma^2}{\sigma^2}\left(\left\|\left(\frac{\partial \widehat{\mathcal{L}}}{\partial \boldsymbol{z}}-\mathbf{1} \mu_g\right)\right\|^2-\left\langle\left(\frac{\partial \widehat{\mathcal{L}}}{\partial \boldsymbol{z}}-\mathbf{1} \mu_g\right), \frac{\hat{\boldsymbol{x}}}{\left\|\hat{\boldsymbol{x}}\right\|}\right\rangle^2\right),\\
		& \le\frac{\gamma^2}{\sigma^2}\left(\left\|\left(\frac{\partial \widehat{\mathcal{L}}}{\partial \boldsymbol{z}}-\mathbf{1} \mu_g\right)\right\|^2-\left\|\frac{\partial \widehat{\mathcal{L}}}{\partial \boldsymbol{z}}-\mathbf{1} \mu_g\right\|^2\cdot \left\|\frac{\hat{\boldsymbol{x}}}{\left\|\hat{\boldsymbol{x}}\right\|}\right\|^2\right). 
	\end{aligned}
\end{equation}
Only $\sigma$ and $\hat{\boldsymbol{x}}$ are different in L1 normalization layers and L2 normalization layers. Then we have: 

\begin{equation}
	\begin{aligned}
		\sigma_1 = |\boldsymbol{x}-\mu|_1,\, \hat{\boldsymbol{x}}_1 = \frac{\bm{x}-\mu}{\sigma_1}, & L1\,normalization \\
		\sigma_2 = |\boldsymbol{x}-\mu|_2,\, \hat{\boldsymbol{x}}_2 = \frac{\bm{x}-\mu}{\sigma_2}, & L2\,normalization. 
	\end{aligned}
\end{equation}
Thus the Lipschitzness constant of L1 normalization $L^1$ and L2 normalization $L^2$ satisfies: 

\begin{equation}
	\frac{L^1}{L^2} \le \frac{\sigma_2^2}{\sigma_1^2} = \frac{|\boldsymbol{x}-\mu|^2_2}{|\boldsymbol{x}-\mu|^2_1}. 
\end{equation}
Moreover, as shown in Figure \ref{l1f}, we record the L1 norm and L2 norm of activations around all training stage. The L1 norm is extremely larger than L2 norm during all training stage, leading to more smooth loss landscape.  

\subsection{Quantization-Toleration of L1 Normalization}\label{qtl1}

\begin{figure}[h]
	\centering 
	\includegraphics[width=1\linewidth, height=0.4\linewidth]{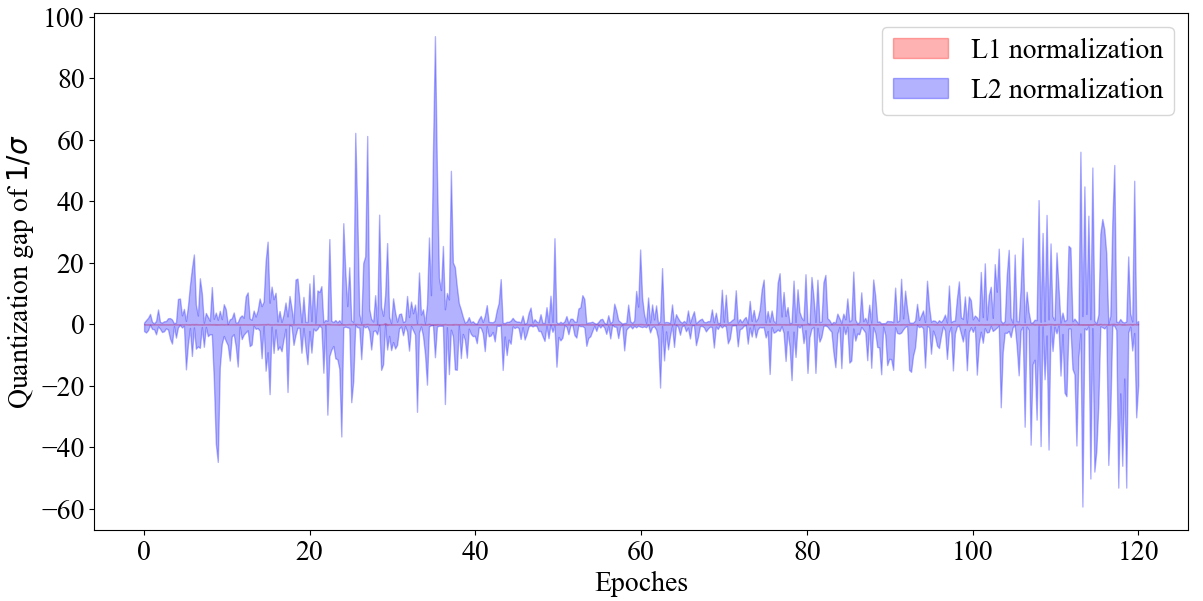}
	\caption{The quantization gap of $\frac{1}{\sigma}$ under L1 normalization and L2 normalization. During the whole training stage, the quantization error of L1 normalization is significantly smaller than L2 normalization. }
	\label{qtl1f}
\end{figure}

In fully-quantized normalization layers, before normalization (Eq. (\ref{normalization})), we first quantize the statistics $\mu$ and $\sigma$: 

\begin{equation}
	\begin{aligned}
		\mu^q = Q(\mu), \sigma^q = Q(\sigma), \, Quantization. 
	\end{aligned}
\end{equation}
\begin{equation}\label{normalization_q}
	\begin{aligned}
		\hat{\boldsymbol{x}} = \frac{\boldsymbol{x}-\mu^q}{\sigma^q}, \, Normalization. 
	\end{aligned}
\end{equation}
In Eq. (\ref{normalization_q}), $\sigma^q$ is the denominator. Same noise imposed on smaller denominator will lead to larger fluctuation in Eq. (\ref{normalization_q}). As shown in Figure \ref{l1f}, $\sigma$ in L2 normalization is significantly smaller than in L2 normalization, which validates the weak quantization-tolerance of L2 normalization layers. Moreover, we visualize the quantization error of $\frac{1}{\sigma}$ under different normalization layers. As shown in Figure \ref{qtl1f}, L1 normalization achieves significantly smaller quantization gap than L2 normalization. 

\section{Detailed implementation of ShiftQuant}\label{shiftMM}

The gradient $\boldsymbol{G _ B} \in \mathbb{R}^{N_B \times C_B}$ is first partitioned into $N _ G$ distinct groups. Each group possesses a unique scaling factor. The $g$-th group $(\boldsymbol{G _ B})^g \in \mathbb{R}^{N_B \times C_{N_g}}$ undergoes quantization according to the following formulation:  

\begin{equation}
	(\boldsymbol{G _ B}) ^ g _ q = round(\frac{(\boldsymbol{G _ B}) ^ g}{\frac{\tau_0}{B}\cdot 2 ^{-g}}) = round(\frac{(\boldsymbol{G _ B})^g}{s ^ {-1} _ {\boldsymbol{G _ B}}\cdot 2 ^ {-g}}), 
\end{equation}

where $\tau_0$ represents the magnitude of $\boldsymbol{G _ B}$. Naturally, we have: 

\begin{equation}
	\begin{aligned}
		(\boldsymbol{G_B}) ^ g \cdot (\boldsymbol{W}^T) ^ g & = [s ^ {-1} _ {\boldsymbol{G _ B}} \cdot 2^{-g} \cdot (\boldsymbol{G_B}) ^ g _ q] \cdot [s _ {\boldsymbol{W}} ^ {-1} \cdot (\boldsymbol{W}^T) ^ g _ q] \\
		& = s ^ {-1} _ {\boldsymbol{G_B}} \cdot s_{\boldsymbol{W}} ^ {-1} \cdot [(\boldsymbol{G_B})^ g _ q \cdot (\boldsymbol{W}^T)^g_q >> g],
	\end{aligned}
\end{equation}

where $\boldsymbol{W} \in \mathbb{R}^{C_A \times C_B}$, and $(\boldsymbol{W}) ^ g _ q \in \mathbb{R}^{C_A \times C _ {N _ g}}$. With aggregating the results from all $N _ G$ groups, we have: 

\begin{equation}
	\begin{aligned}
		\boldsymbol{G _ A} &= \sum_{g=1} ^ {N _ G}{{(\boldsymbol{G} _ B)} ^ g} \cdot {{(\boldsymbol{W} ^ T)} ^ g} \\
		&= s ^ {-1} _ {\boldsymbol{G} _ B} \cdot s ^ {-1} _ {\boldsymbol{W}} \sum_{g=1} ^ {N _ G}[{{(\boldsymbol{G} _ B)} ^ g} _ q \cdot {{(\boldsymbol{W} ^ T)} ^ g} _ q >> g],
	\end{aligned}
\end{equation}

In the code implementation, we substitute the right shift operation with a left shift operation. This modification ensures that the outcome is perfectly congruent with the original dot product computation: 

\begin{equation}
	\begin{aligned}
		 & s^{-1}_{\boldsymbol{G}_B} \cdot s^{-1}_{\boldsymbol{W}} \cdot 2 ^ {-N_G} \sum_{g=1} ^ {N _ G}[{{(\boldsymbol{G} _ B)} ^ g} _ q \cdot {{(\boldsymbol{W} ^ T)} ^ g} _ q << (N_g - g)] \\
		=& \sum_{g=1}^{N_ G} s ^ {-1} _ {\boldsymbol{G} _ B} \cdot s ^ {-1} _ {\boldsymbol{W}} \cdot 2 ^ {-N_G} [{{(\boldsymbol{G} _ B)}^g}_q \cdot {{(\boldsymbol{W} ^ T)} ^ g} _ q << (N_g - g)] \\
		=& \sum_{g=1} ^ {N _ G} s ^ {-1} _ {\boldsymbol{G} _ B} \cdot {{(\boldsymbol{G} _ B)} ^ g} _ q \cdot (2 ^ {-N_G} << (N_g - g)) \cdot s ^ {-1} _ {\boldsymbol{W}} \cdot {{(\boldsymbol{W} ^ T)} ^ g} _ q \\
		=&  \sum_{g=1} ^ {N _ G} s^{-1} _ {\boldsymbol{G}_B} \cdot {{(\boldsymbol{G} _ B)} ^ g} _ q \cdot 2 ^ {-g} \cdot s^{-1}_{\boldsymbol{W}} \cdot {{(\boldsymbol{W}^T)}^g}_q \\
		=& \sum_{g=1} ^ {N_G} {{(\boldsymbol{G}_B)}^g} \cdot {{(\boldsymbol{W} ^ T)} ^ g} \\
		=& \boldsymbol{G_A}. 
	\end{aligned}
\end{equation}

\section{Implementation of DSP Reusing on FPGA}\label{dspreusing}

\begin{figure}[h]
	\centering 
	\includegraphics[width=1\linewidth, height=0.28\linewidth]{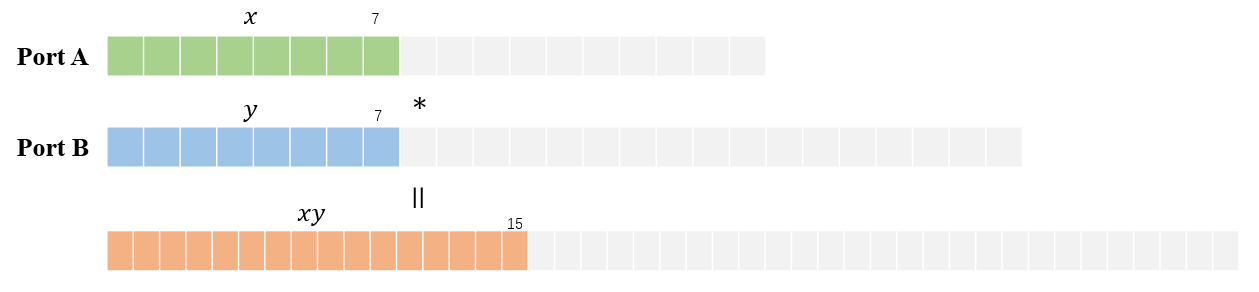}
	\caption{Direct utilization of DSP on 8-bit multiplication. Unused bits is colored in gray. A lot of bitwidth is wasted. }
	\label{dsp1}
\end{figure}

In the Xilinx ZC706 board, the interface for the multiplier within the FPGA's DSP module (Xilinx DSP48E1) is configured for 25-bit and 18-bit inputs. In low-bitwidth computations, the direct utilization of DSPs leads to the wastage of bit width (as shown in Figure \ref{dsp1}). DSP reusing addresses this issue by packing multiple low-bitwidth data into a single data unit sized to the bit width. This approach enables the execution of multiple multiplications in a single computation cycle. 

\begin{figure}[h]
	\centering 
	\includegraphics[width=1\linewidth, height=0.28\linewidth]{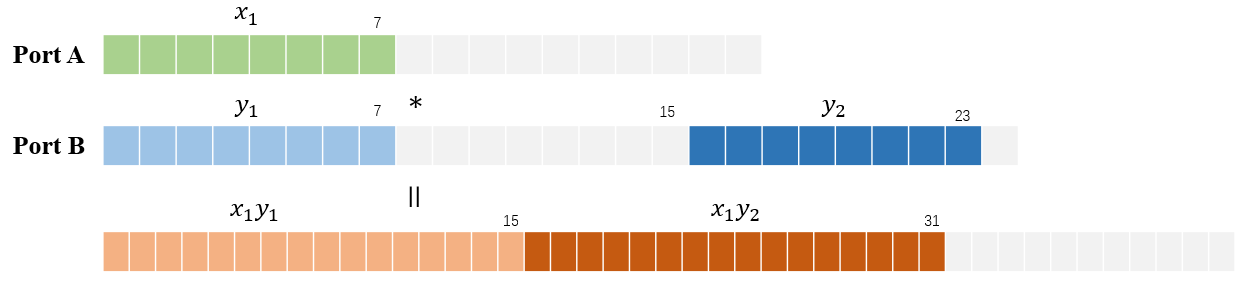}
	\caption{DSP reusing on 8-bit multiplication. One calculation can realize two multiplications. }
	\label{dsp2}
\end{figure}

\begin{figure}[h]
	\centering 
	\includegraphics[width=1\linewidth, height=0.28\linewidth]{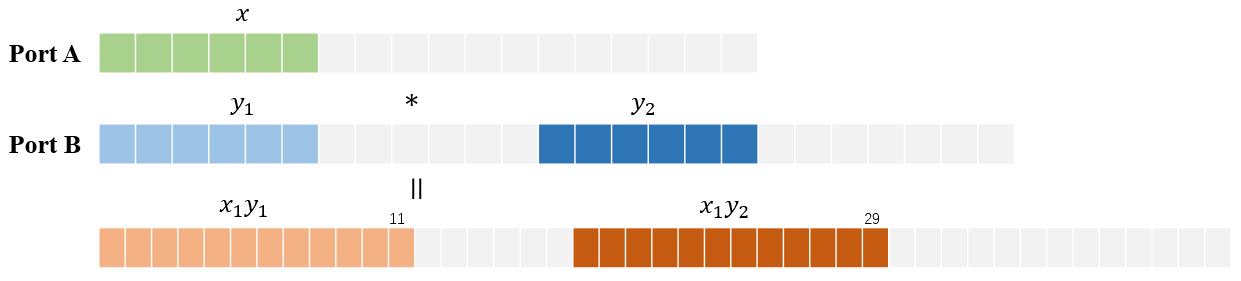}
	\caption{DSP reusing on 6-bit multiplication. One calculation can realize two multiplications. }
	\label{dsp3}
\end{figure}

As shown in Figure \ref{dsp2} and \ref{dsp3}, for 8-bit and 6-bit multiplications, the Xilinx DSP48E1 block can achieve a maximum of dual multiplexing, allowing for the execution of two multiplications in a single operation. This capability is also the reason why the DSP resource consumption for both 6-bit and 8-bit implementations is the same, as indicated in Table \ref{fpga}. 

\begin{figure}[h]
	\centering 
	\includegraphics[width=1\linewidth, height=0.28\linewidth]{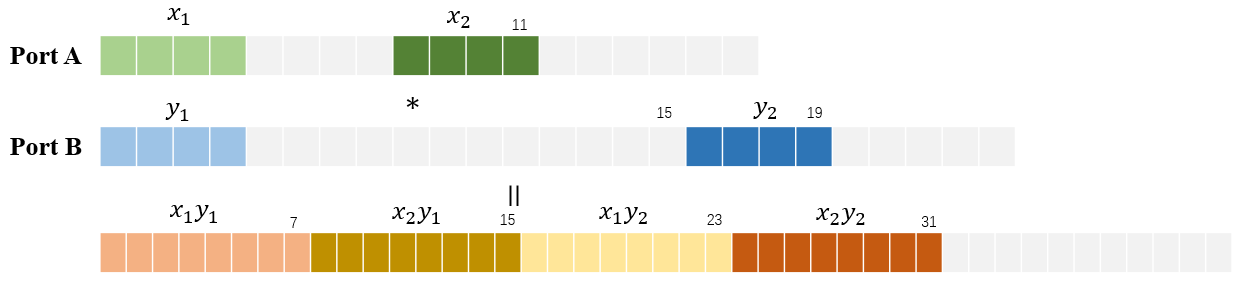}
	\caption{DSP reusing on 4-bit multiplication. One calculation can realize four multiplications. }
	\label{dsp4}
\end{figure}

As shown in Figure \ref{dsp4}, the DSP48E1 block is capable of simultaneously executing four 4-bit multiplications. This aligns with the data presented in Table \ref{fpga}, where the DSP resource consumption for the 4-bit implementation is half that of the 6-bit and 8-bit implementations.

\end{document}